\documentclass{article}

\usepackage[preprint]{style/neurips_2025}

\usepackage[utf8]{inputenc}
\usepackage[T1]{fontenc}
\usepackage{url}
\usepackage{booktabs}
\usepackage{amsmath,amsfonts,amssymb,mathtools}
\usepackage{bbold}
\usepackage{wrapfig}
\usepackage{graphicx}
\usepackage{caption}
\usepackage{subcaption}
\usepackage{tabularx}
\usepackage{float}
\usepackage{xcolor}
\usepackage{multirow}
\usepackage{threeparttable}
\usepackage{enumitem}
\usepackage{dsfont}
\usepackage{bm}
\usepackage[ruled,vlined,linesnumbered]{algorithm2e}
\usepackage{hyperref}

\graphicspath{{figures/}}

\usepackage{xcolor}
\usepackage{hyperref}

\title{LLM-TabLogic: Preserving Inter-Column Logical Relationships in Synthetic Tabular Data via Prompt-Guided Latent Diffusion}

\author{%
  Yunbo Long\thanks{Corresponding author: \texttt{yl892@cam.ac.uk}} \\
  Department of Engineering\\
  University of Cambridge\\
  Cambridge, United Kingdom\\
  \texttt{yl892@cam.ac.uk} \\
  \And
  Liming Xu \\
  Department of Engineering\\
  University of Cambridge\\
  Cambridge, United Kingdom\\
  \texttt{lx249@cam.ac.uk} \\
  \And
  Alexandra Brintrup \\
  Department of Engineering, University of Cambridge\\
  The Alan Turing Institute\\
  Cambridge / London, United Kingdom\\
  \texttt{ab702@cam.ac.uk} \\
}

\begin{document}

\maketitle

\begin{abstract}
Synthetic tabular data are increasingly being used to replace real data, serving as an effective solution that simultaneously protects privacy and addresses data scarcity. However, in addition to preserving global statistical properties, synthetic datasets must also maintain domain-specific logical consistency—especially in complex systems like supply chains, where fields such as shipment dates, locations, and product categories must remain logically consistent for real-world usability. Existing generative models often overlook these inter-coloumn relationships, leading to unreliable synthetic tabular data in real-world application.
To address these challenges, we propose LLM-TabLogic, a novel approach that leverages Large Language Model (LLM) reasoning to capture and compress
the complex logical and relationships among tabular columns, while these conditional information are passed into a Score-based Diffusion for data generation in latent space.
Through extensive experiments on real-world industrial datasets, we evaluate LLM-TabLogic for column reasoning and data generation, comparing it with five baselines including SMOTE and state-of-the-art generative models.
Our results show that LLM-TabLogic demonstrates strong generalization in logical inference, achieving over 90\% accuracy on unseen tables. And our method outperforms all baselines in data generation by fully preserves inter-column relationships while maintaining the best balance between data fidelity, utility, and privacy.
This study presents the first method to effectively preserve inter-column relationships in synthetic tabular data generation without requiring domain knowledge, offering new insights for creating logically consistent real-world tabular data. The code is available at
\href{https://github.com/Yunbo-max/TabKG}
{\textcolor{red}{https://github.com/Yunbo-max/TabKG}}.
\end{abstract}


\section{Introduction}
\label{sec:introduction}
Synthetic data is {\it artificially} generated using original data and a model trained to replicate the characteristics and structure of the original dataset \citep{jordon2022synthetic}.
It is widely recognized as a solution for enabling secure and privacy-preserved data sharing while also addressing challenges associated with limited access to real data \citep{long2025leveraging}.
Given the critical need for privacy-preserved data exchange and the scarcity of real data, synthetic data, particularly synthetic {\it tabular} data, has the potential for widespread application in industrial scenarios, such as supply chain analytics, operations optimization, and digital twin simulation.

Tabular data, often structured in a table with rows and columns and stored in relational databases \citep{vieira2020supply}, is {\it ubiquitous} \citep{van2024tabular} across various industries---ranging from medical records in healthcare and transaction data in finance to sales data in retail to inventory data in warehousing.
Such data is often carefully designed and maintained by domain experts, with data schemes or structures that are domain-specific and where columns are {\it interdependent}.
Compared to unstructured data modalities such as image, video, and text, tabular data is structurally simpler yet semantically more complex, featuring well-defined columns, rich inter-column interactions, high dimensionality and heterogeneity, as well as domain-specific context and complexity.
Thus, synthetic {\it tabular} data generation is challenging and remains underexplored, with its development significantly lags behind these unstructured data modalities \citep{van2024tabular}.

Nevertheless, many methods \citep{chawla2002smote,xu2019modeling,kotelnikov2023tabddpm,borisov2022language}---mainly adapted from models designed for generating other data modalities---have been proposed in literature to generate tabular data.
One of earliest such approaches is the Synthetic Minority Over-sampling TEchnique (SMOTE), originally proposed by \citet{chawla2002smote} to address dataset imbalance problem. 
As a data augmentation method, this interpolation-based method has recently been found effective in generating tabular data \citep{wang2024challenges}.
However, despite its promising performance on several benchmark datasets, as reported in \citet{long2025evaluating}, SMOTE inherently struggles to preserve data privacy effectively and capture complex column dependencies.
To overcome these limitations, deep generative models—leveraging their success in image, text, and audio synthesis—have been increasingly explored for tabular data generation.
For instance, \citet{xu2019modeling} proposed CTGAN, a generative adversarial network (GAN)-based method specifically designed for tabular data.
CTGAN learns the distribution of tabular data and generates synthetic rows from the learned representation. 
Although CTGAN improves privacy preservation \citep{zhao2024ctab}, it struggles to capture the interdependencies between categorical and continuous columns \citep{zhang2023mixed}. 
More recently, diffusion model-based methods such as TabDDPM \citep{kotelnikov2023tabddpm} and TabSyn \citep{zhang2023mixed} have been introduced to synthesize tabular data by learning the denoising process in {\it latent} space. 
These approaches demonstrate superior performance compared to CTGAN, particularly in preserving intricate tabular structures. 
Furthermore, \citet{borisov2022language} proposed GReaT, a large language model (LLM)-based approach for tabular data generation.
GReaT transforms tabular data into textual representations and fine-tunes generative models to produce synthetic data, enabling the learning of underlying distributions without requiring complex encoding and decoding processes.

While these generative methods demonstrate effectiveness in generating tabular data for machine learning tasks---particularly for classification and prediction---they often overlook or fail to explicitly preserve inter-column relationship in the generated data, a property crucial for real-world applications.
Inter-column relationships are often thus governed by domain-specific constraints \citep{long2025evaluating}, such as the temporal requirement that delivery times must occur after ordering times in logistics or the mathematical constraint that the product of unit price and quantity must equal the total amount in procurement.
Due to their failure to preserve inter-column relationships, existing generative methods---primarily optimized for facilitating machine learning tasks---may produce unrealistic tabular data that lack real-world utility, limiting their applicability in simulation and informed decision-making.
Moreover, despite the critical importance of inter-column relationships in industrial tabular data generation, few prior studies have explicitly focused on this aspect.

In this paper, we thus aim to bridge this gap.
{\it First}, we propose an evaluation framework to assess the effectiveness of synthetic tabular data in real-world scenarios. 
{\it Next}, we introduce LLM-TabLogic, a LLM-based synthetic tabular data generation method designed to preserve both privacy and inter-column relationships.
This approach leverages LLMs to compress tabular data by inferring relationships and employs a score-based diffusion model to train and sample synthetic data in the latent space.
{\it Finally}, we conduct experiments to comprehensively evaluate the proposed method and compare it with both traditional and state-of-the-art approaches.
Experimental results on two real-world industrial datasets demonstrate that LLM-TabLogic outperforms existing methods under the proposed evaluation framework across three key dimensions: data fidelity, data utility, and data privacy, achieving a better trade-off between data fidelity and privacy preservation.
By effectively preserving the complex inter-column dependencies, LLM-TabLogic is particularly well-suited for real-world applications beyond traditional machine learning tasks.
It is the first method, to the best of our knowledge, to explicitly focus on capturing complex inter-column relationships in tabular data generation. 
By prioritizing relational consistency within tables, LLM-TabLogic sets a new benchmark for creating realistic and reliable synthetic datasets.

The rest of this paper is structured as follows. \autoref{sec:related_work} reviews related work.
\autoref{sec:LLM-TabLogic} details the proposed approach, LLM-TabLogic.
\autoref{sec:setting} presents experimental settings, including datasets, baselines, along with the evaluation framework.
\autoref{sec:results} presents experimental results. 
\autoref{sec:discussion} discusses the limitation and implications of the proposed approach.
Finally, \autoref{sec:conclusion} concludes the paper and summarizes future work.

\section{Related Work}
\label{sec:related_work}
This section reviews related work, including the preliminaries, the complexities of tabular data, and relevant methods for generating synthetic tabular data.

\subsection{Tabular Data and Its Inherent Complexity}
Tabular data, typically structured in a table with rows and columns and stored in relational databases, is the dominant modality \citep{vieira2020supply,van2024tabular} in many fields, such as healthcare \citep{nik2023generation,hernandez2022synthetic}, finance \citep{sattarov2023findiff}, and supply chains \citep{long2025leveraging}.
While modern generative models have achieved human-like performance in synthesizing other data modalities, such as text, images, and audio, as demonstrated by ChatGPT \citep{wu2023brief}, DALL-E 2 \citep{lin2023standing}, and Suno \citep{yu2024suno}, the development of synthetic tabular data remains in its early stage and significantly lags behind these advancements.

This disparity is largely due to the complexity inherent in tabular data, which differs fundamentally from other data modalities. 
Unlike unstructured data modalities---such as text and images---that are typically one- or two-dimensional and follow well-defined spatial and temporal distributions, tabular data, though structured, is more complex and often lacks clear structural patterns, featuring both categorical and continuous columns with rich inter-column dependencies.
Moreover, the lack of a unified format across features, each governed by its own distinct distribution, further complicates the generation of high-quality synthetic tabular data \citep{shan2019crowdsourcing,wang2024harmonic}.
This heterogeneity makes it challenging to capture and preserve the underlying inter-column relationships columns while ensuring data fidelity and utility.
These relationships \citep{long2025evaluating}, whether mathematical, temporally sequential, or geographically hierarchical, are often defined based on domain-specific knowledge and differ significantly from one domain to another.
As a result, although effective for other data modalities, current state-of-the-art methods aimed at enhancing machine learning tasks struggle to preserve inter-column relationships---a critical aspect for ensuring the utility of synthetic tabular data in real-world applications.
This highlightthe need for more focused research in this aspect.

\subsection{Evaluation Framework}

Currently, most research on tabular data generation methods is validated using low-dimensional datasets, such as Adult Income \citep{kohavi1996scaling} and California Housing \citep{pace1997sparse}. In contrast, fewer studies explore large-scale datasets like HIGGS \citep{baldi2014searching} and Covertype \citep{asuncion2007uci}. However, even these large-scale datasets predominantly consist of numerical attributes and lack complex inter-column relationships, particularly between categorical and numerical data. This limits their ability to represent real-world scenarios where logical consistency and dependencies are critical.
And recent advancements in synthetic tabular data generation have been evaluated on real-world industrial datasets, particularly in finance and healthcare. For example, studies have used datasets such as Payments and Credit \citep{sattarov2023findiff} and MIMIC-III \citep{johnson2016mimic}. These datasets inherently involve complex inter-column relationships, such as transaction amounts linked to account balances in financial data or patient diagnoses influencing treatment plans in healthcare. Despite the presence of these relationships, evaluations in these studies have largely overlooked their preservation. Instead, the focus has been on traditional metrics like machine learning utility and statistical alignment \citep{papadaki2024exploring, van2024latable}.
Specifically, the current evaluation of data fidelity primarily focuses on capturing pairwise column correlations and multivariate dependencies \citep{margeloiu2024tabebm}. While these metrics provide insights into statistical relationships, they fail to guarantee the consistency or preservation of logical relationships within multivariate distributions \citep{margeloiu2024tabebm}. For example, in industrial tabular data from supply chains, complex inter-column relationships are often present, such as order dates being logically linked to delivery dates or total sales prices calculated as a function of price, quantity, and discounts \citep{suh2024timeautodiff, borisov2022language}. 
Moreover, some logical relationships such as hierarchical consistency (e.g., city and country) are not unified. 
And conditional dependencies between those features may change depending on the specific combination of values in each row. This variability adds another layer of complexity to accurately modeling and evaluating these relationships. Recognizing this challenge, \citep{long2025evaluating} proposed the first evaluation framework specifically designed for inter-column relationship preservation. Their work highlights the need for advanced generative models capable of accurately modeling these intricate, row-specific dependencies.

\subsection{Tabular Data Generation Approaches}

One of the earliest effective methods for generating tabular data is the Synthetic Minority Over-sampling TEchnique (SMOTE), originally proposed by \citet{chawla2002smote} to address dataset imbalance. 
As a data augmentation method, this interpolation-based method has recently been found effective in generating tabular data \citep{wang2024challenges}.
Despite its simplicity, SMOTE achieves performance comparable to many advanced generative models on several benchmark datasets, as reported by \citet{long2025evaluating}.
Mathematically, SMOTE can be defined as:
\begin{equation}
\mathbf{x}_{new} = \mathbf{x}_i + \lambda (\mathbf{x}_i - \mathbf{x}_j)
\end{equation}
where \( \mathbf{x}_i \) and \( \mathbf{x}_j \) are two randomly chosen instances from the minority class, and \( \lambda \) is a random number between 0 and 1, controlling the distance between \( \mathbf{x}_i \) and \( \mathbf{x}_j \). 
However, since SMOTE generates new data points by linearly interpolating among neighboring data points for each feature, it inherently struggles to capture complex relationships and dependencies among diverse columns. 
Therefore, deep generative models have been introduced for tabular data generation. 
\citet{xu2019modeling} proposed TVAE, a Variational Autoencoder-based method, for generating tabular data.
They further presented Conditional Generative Adversarial Networks (CTGAN), which achieved better performance than TVAE in terms of data accuracy.
CTGAN transforms input categorical data into numerical formats using techniques such as one-hot encoding \citep{kim2022stasy} and analog bits encoding \citep{zheng2022diffusion}. 
The discriminator \( D \) is trained to maximize the following objective:
\begin{equation}
\mathcal{L}_D = - \mathbb{E}_{\mathbf{x} \sim p_{data}}[\log D(\mathbf{x})] - \mathbb{E}_{\mathbf{z} \sim p_z}[\log (1 - D(G(\mathbf{z})))].
\label{eq:discrimiator}
\end{equation}
The discriminator \( D \) in Eq. \ref{eq:discrimiator} is trained to distinguish between real data \( \mathbf{x} \) and synthetic data \( G(\mathbf{z}) \), where \( \mathbf{z} \) is drawn from a noise distribution \( p_z \). This process focuses on optimizing the decision boundary between real and fake samples, capturing global patterns and correlations. However, it does not explicitly model discrete logical relationships, such as conditional dependencies (e.g., "if city = London, then country = UK"). These logical relationships require higher-level semantic understanding, which the discriminator, by focusing on broader correlations, cannot guarantee in the generated data.
Inspired by the remarkable success of diffusion models in in image \citep{ho2020denoising} and audio generation \citep{huang2023make}, diffusion model-based methods have been proposed for tabular data generation, such as TabDDPM \citep{kotelnikov2023tabddpm} and TabSyn \citep{zhang2023mixed}.
Unlike TVAE and CTGAN, which rely on adversarial training and may struggle with issues like mode collapse or imbalanced data, diffusion model-based methods use a gradual denoising process, ensuring smoother and more stable learning of the tabular data distribution. 
In the backward process (Equation~\ref{eq:diffusion_sample}), the transition from the original data $\mathbf{x}_0$ to the noisy data $\mathbf{x}_t$ is modeled as:
\begin{equation}
    q(\mathbf{x}_t | \mathbf{x}_0) = \mathcal{N}(\mathbf{x}_t; \sqrt{\bar{\alpha}_t} \mathbf{x}_0, (1 - \bar{\alpha}_t) \mathbf{I}),
\label{eq:diffusion_sample}
\end{equation}
where $\bar{\alpha}_t = \prod_{i=1}^t \alpha_i$ represents the cumulative noise scaling factor at step $t$. The backward process aims to denoise $\mathbf{x}_t$ to recover $\mathbf{x}_0$. 
For tabular data, even when embedded in a continuous space $\mathcal{X}$, feature relationships remain governed by discrete logical constraints rather than spatial correlations. Unlike natural images where neighboring pixels exhibit smooth variations \citep{xu2019modeling}, tabular embeddings may represent categorical variables or discontinuous relationships (e.g., $\text{age} < 18 \Rightarrow \text{is\_adult}=0$). This violates the local smoothness prior that makes diffusion models effective for images, requiring alternative approaches for tabular data denoising.
Recent advances in pretrained language models (PLMs) have inspired LLM-based tabular data generation methods including GReaT \citep{borisov2022language}, REaLTabFormer \citep{solatorio2023realtabformer}, and Tabula \citep{zhao2023tabula}. These approaches tokenize tabular data and fine-tune PLMs using the objective function:
\begin{equation}
    L(\theta; T) = - \sum_{t_i \in T} \sum_{j=1}^{m} \log P(t_{i,j} | \Pi(t_i, k)_{1:j-1}),
    \label{eq:loss_function}
\end{equation}
where $P(t_{i,j} | \Pi(t_i, k)_{1:j-1})$ represents the conditional probability of token prediction given preceding context. While this formulation effectively preserves inter-column logical relationships through autoregressive modeling, LLM-based methods face several challenges compared to GANs and diffusion models. The sequential nature of PLMs leads to substantially longer training and generation times on tabular data, while introducing sensitivity to feature ordering that can significantly impact model effectiveness \citep{xu2024llms}. Furthermore, the tokenization process and autoregressive objective struggle to accurately model continuous numerical distributions that frequently appear in tabular datasets, particularly for precise value generation in real-valued features\citep{}.

While existing methods demonstrate effectiveness in privacy preservation and in maintaining sequential inter-column relationships to some extent, they fall short in accurately preserving more complex inter-column dependencies in the generated synthetic data.
We thus address this gap in the paper, proposing a LLM-based tabular data generation approach---LLM-TabLogic---which will be detailed in the following section.

\section{LLM-TabLogic Approach}
\label{sec:LLM-TabLogic}
\begin{figure*}[t!]
    \centering
    \includegraphics[width=1\textwidth]{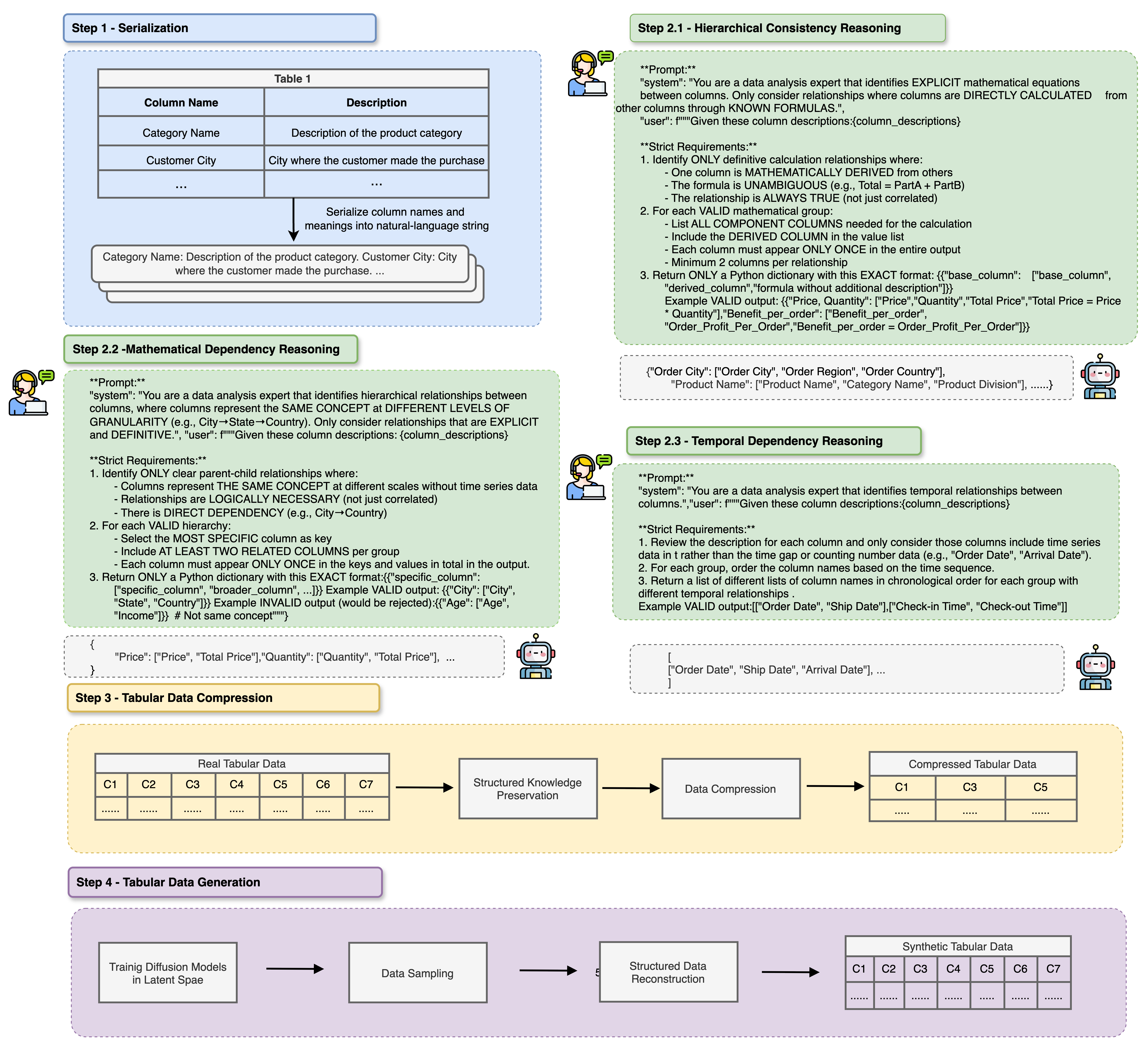}
    \caption{Overview of the workflow of the LLM-TabLogic approach.}
    \label{fig:LLM-TabLogic}
\end{figure*}

Recently, LLMs such as ChatGPT \citep{brown2020language}, LLaMa \citep{touvron2023llama}, and Deepseek \citep{guo2025deepseek} have been demonstrated remarkable success in natural language processing and beyond.
These models enable the automation of complex data processing tasks and significantly enhance the semantic understanding of text, including tabular data. 
For instance, \citet{li2024optimization} demonstrated their ability to summarize tabular data from reports, while \citet{liu2023rethinking} showcased their potential for textual and symbolic reasoning on tables, facilitating deeper analysis and complex query resolution. 
These advancements highlight LLMs' ability to bridge natural language understanding and structured data analysis.
However, pretraining LLMs on large volumes of tabular text data \citep{hegselmann2023tabllm} demands significant computational resources and time.
Additionally, tabular data lacks the structured context LLMs rely on, making them less effective for {\it generation} tasks compared to their strengths in text understanding.
In contrast, deep generative models, such as diffusion models, excel at capturing complex patterns in tabular data by learning latent representations. 
These models efficiently model the joint distribution of features, enabling high-quality synthetic data generation. 
While deep generative models are adept at replicating statistical correlations, LLMs offer unique advantages in understanding semantic relationships and preserving inter-column dependencies.
Therefore, we propose LLM-TabLogic, a novel framework that addresses tabular data synthesis through two key innovations: First, we develop prompt-based methods to reason about and compress complex inter-column relationships while preserving their logical structure. Second, we demonstrate how these extracted conditional patterns can effectively guide generative models in the latent space to produce logically consistent synthetic data.
The workflow of LLM-TabLogic is illustrated in \autoref{fig:LLM-TabLogic}, with each step detailed in the following sections.

\subsection{Serialization}
The first step of LLM-TabLogic is serialization, as shown in \autoref{fig:LLM-TabLogic}.
Let $X = [x_1, x_2, \dots, x_n]$ represent the tabular dataset, where $x_i$ corresponds to the values of the $i$-th column, and $C = [c_1, c_2, \dots, c_n]$ denote the corresponding column names. 
Additionally, let $D = [d_1, d_2, \dots, d_n]$ denotes the column descriptions, where $d_i$ describes the semantics or metadata of the column $c_i$; $Y$ represents the target data (dependent variable). 
In this step, the column names and descriptions are serialized into natural language using the following mapping function:  
\begin{equation}
    F:(c_j, d_j) \mapsto  c_j \text{ : } d_j, j = 1, 2, \dots, n
\end{equation}  
where \( c_j \) represents a column name and \( d_j \) its corresponding description.
Once the dat is serialized, the next task is to use the serialized data, \( F(c_j, d_j) \) as input to a LLM to infer logical groupings among columns.
The relationships between the columns are determined based on these serialized descriptions.
The inferred relationships, combined with expert-designed domain knowledge, is used to create a general prompt \( p \) and task-specific instructions \( i \), forming the input for the LLM.
This input is represented as \( (F(c_j, d_j), p, i) \), where the LLM utilizes this information to identify meaningful relationships among multiple columns.
Finally, the LLM generates lists of column names that are logically connected based on the serialized data and instructions provided.

\subsection{LLM Reasoning}\label{sec:llm_reasoning}
The second step is LLM reasoning. 
We illustrate examples of prompts and instruction inputs for key logical relationships in tabular data in \autoref{fig:LLM-TabLogic}, including hierarchical consistency and logical dependencies, such as mathematical and temporal relationships.
The output of LLM reasoning is structured as dictionaries and lists to represent inter-column relationships.  Specifically, hierarchical consistency is represented as a set of mappings:  
\begin{equation}
    \mathcal{R}^\mathcal{H} = \Bigl\{ c^h_{1,j} : \{ c^h_{2,j}, c^h_{3,j}, \dots, c^h_{n,j} \} \Bigr\},
\end{equation}
where \( c^h_{1,j} \) is the most granular column in the semantic hierarchy of the \(j \) th row, and \( \{ c^h_{2,j}, c^h_{3,j}, \dots, c^h_{n,j} \} \) are other columns within the same hierarchical group \( h \). Here, \( h \) can also take on values such as \( 1, 2, 3, \ldots, k \), representing different hierarchical groups. Besides, \( n \) represents the number of additional columns that belong to the same hierarchical group.  
Mathematical dependencies are defined as a mapping between independent and derived columns:  
\begin{equation}
    \mathcal{R}^\mathcal{M} = \Bigl\{ \{ f^m_{i_1}, f^m_{i_2}, \dots, f^m_{i_n} \} : \{ f^m_{d_1}, f^m_{d_2}, \dots, f^m_{d_q} \} \Bigr\},
\end{equation}
where \( \{ f^m_{i_1}, f^m_{i_2}, \dots, f^m_{i_n} \} \) are independent values of n columns, and \( \{ f^m_{d_1}, f^m_{d_2}, \dots, f^m_{d_q} \} \) are derived columns that depend on them. The values of \( q \) and \( n \) indicate the number of columns in the respective sets. Additionally, \( t \in \{ 1, 2, 3, \ldots, k \} \), where \( m \) indicates the mathematical dependency group.
Temporal dependencies are represented as an ordered sequence of columns:  
\begin{equation}
    \mathcal{R}^\mathcal{T} = \Bigl[ f^t_{1}, f^t_{2}, \dots, f^2_{n} \Bigr]
\end{equation}  
where \( f^t_{1}, f^t_{2}, \dots, f^2_{n} \) are columns arranged in chronological order, ensuring that temporal relationships between features are maintained. Additionally, \( t \in \{ 1, 2, 3, \ldots, k \} \), where \( t \) indicates the  dependency group. The parameter \( n \) denotes the total number of columns in the ordered sequence.

\subsection{Tabular Data Compression}

Many real-world datasets contain deterministic relationships---fixed rules where one variable is perfectly predictable from others, such as functional dependencies (\textit{City} from \textit{ZIP Code}), arithmetic constraints ($\textit{Total} = \sum \textit{Parts}$), or temporal sequences ($\textit{Departure\_Date} < \textit{Arrival\_Date}$). These relationships introduce no uncertainty ($P(X_i | X_{-i}) = 1$) and thus provide no meaningful signal for probabilistic models like diffusion or drift-based approaches to learn. 
Instead of wasting model capacity on reconstructing rigid, rule-based mappings, we can preprocess these dependencies out---either by removing existing derived variables or by replacing them with newly derived values. 
Recent advances have demonstrated large language models' (LLMs) capabilities in tabular data processing, from summarization \citep{li2024optimization} to symbolic reasoning \citep{liu2023rethinking}. Building on these, we employ an LLM as a \textit{relational conditioner} that compresses its logical relationships into a structured latent representation $\mathcal{R}$. This representation captures three fundamental dependencies: \textit{hierarchical} ($\mathcal{R}^\mathcal{H} = \{c^h_1: \{c^h_2,...\}\}$) through leaf nodes $X_\ell$ with invertible ancestor reconstruction $X_a = \phi_a(X_\ell)$; \textit{mathematical} ($\mathcal{R}^\mathcal{M} = \{\text{independent}\} : \{\text{derived}\}$) via minimal sufficient statistics $S=X_j$ and Dirac delta constraints $\delta(X_k-\psi(X_j))$; and \textit{temporal} ($\mathcal{R}^\mathcal{T} = [f^t_1,...,f^t_n]$) using initial states $X_1$ with Markovian increments $\{\Delta_t\}$. The reconstruction $\mathcal{D} = \mathcal{C}^{-1}$ preserves pairwise distributions $p_{\mathbf{X}}(x_j,x_k)$ through: (i) exact invertibility of $\phi_a$ and $\psi$ ensuring $p(\hat{x}_j,\hat{x}_k) = p(x_j,x_k)$, and (ii) Markov consistency via Chapman-Kolmogorov equations \citep{wikichapmankolmogorov}. 
And those Learned restoration functions $\{f_{decom}\}$ are subsequently used to condition our latent-space diffusion model during data generation.

\subsection{Compressed Data Generation}
\begin{figure*}[t]
    \centering
    \includegraphics[width=0.80\textwidth]{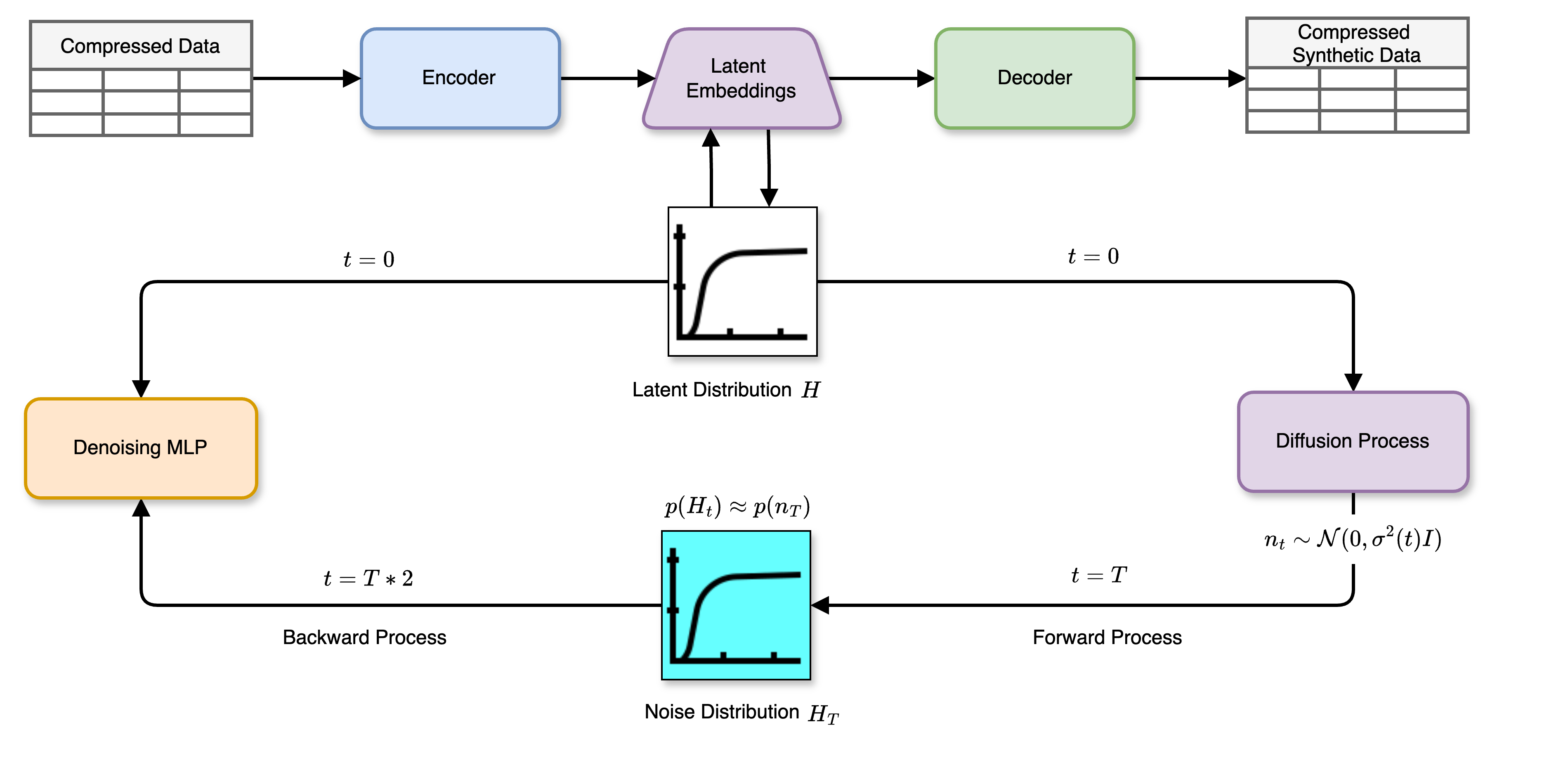}
    \caption{Illustration of compressed data generation via score-based diffusion in the latent space.}
    \label{fig:ddpm}
\end{figure*}

The compressed data $\mathcal{X}_{\text{c}}$ is used to train a score-based diffusion model for generating synthetic tabular data $\mathcal{X}_{\text{c}}^{'}
$. 
Following the preprocessing methods in \citet{zhang2023mixed}, we integrate a tokenizer and a Transformer-based Variational Autoencoder (VAE) into a unified encoder. Specifically, we first tokenize $\mathcal{X}_{\text{c}}$ to represent mixed data types (numerical and categorical) within a unified dense vector space. The tokenized data is then mapped into a latent space via the Transformer VAE, where the encoder outputs the mean and log variance of the latent distribution. 
By reparameterizing the latent distribution, we derive the latent embedding, denoted as $\mathbf{H}_0$, which serves as the initial representation of the data in the latent space. 
The diffusion model is then employed to learn the underlying distribution of $\mathbf{H}_0$ by progressively transforming it through a forward process and subsequently recovering it via a reverse process.
To model the latent distribution $p(\mathbf{H}_0)$, we employ a score-based diffusion framework \citep{karras2022elucidating}, which consists of a forward stochastic perturbation process and a reverse generative process. The forward process incrementally introduces Gaussian noise to the initial latent embedding $\mathbf{H}_0$, transforming it into a noisy latent state $\mathbf{H}_T$ over time. This evolution is described by the following equation:
\begin{equation}
\mathbf{H}_t = \mathbf{H}_0 + \alpha(t) \boldsymbol{\eta}, \quad \boldsymbol{\eta} \sim \mathcal{N}(0, \mathbf{I})
\end{equation}
where $\alpha(t)$ represents the noise scaling function at time $t$, determining the intensity of the perturbation applied to the latent representation.
The reverse process seeks to reconstruct the original latent state by iteratively denoising $\mathbf{H}_t$. This is governed by the following stochastic differential equation:
\begin{equation}
d\mathbf{H}_t = -2 \alpha'(t) \alpha(t) \nabla_{\mathbf{H}_t} \log p(\mathbf{H}_t) \, dt + \sqrt{2 \alpha'(t) \alpha(t)} \, d\mathbf{W}_t
\end{equation}
where $\nabla_{\mathbf{H}_t} \log p(\mathbf{H}_t)$ is the learned score function, which estimates the gradient of the log-density and guides the denoising process.
And $\alpha'(t)$ represents the time derivative of the noise scaling function $\alpha(t)$, meaning it quantifies how the noise level changes over time. 
The term $\mathbf{W}_t$ denotes the standard Wiener process, introducing controlled stochasticity into the reverse dynamics.
To achieve accurate reconstruction, the model is trained using denoising score matching. Specifically, a neural network $\boldsymbol{\eta}_\phi(\mathbf{H}_t, t)$ is optimized to approximate the noise added during the forward process. The training objective is to minimize the following loss function \citep{karras2022elucidating}:
\begin{equation}
\mathcal{L} = \mathbb{E}_{\mathbf{H}_0 \sim p(\mathbf{H}_0)} \mathbb{E}_{t \sim p(t)} \mathbb{E}_{\boldsymbol{\eta} \sim \mathcal{N}(0, \mathbf{I})} \| \boldsymbol{\eta}_\phi(\mathbf{H}_t, t) - \boldsymbol{\eta} \|^2_2.
\end{equation}
Once the diffusion model is trained, synthetic latent embeddings $\mathbf{H}'$ can be sampled by applying the learned reverse process. These embeddings are subsequently decoded to reconstruct synthetic data representations that preserve the statistical and structural properties of the original dataset. This generative mechanism ensures that the synthetic data $\mathcal{X}_{\text{c}}'$ remains faithful to the distribution of the original compressed data $\mathcal{X}_{\text{c}}$, facilitating high-quality synthetic data generation.

\subsection{Synthetic Data Decompression}

The compressed synthetic data $\mathcal{X}_{\text{c}}^{'}
$ is decompressed into a reconstructed dataset $\mathcal{X}^{'}
$ using relationship-guided mappings $\mathcal{R}_h$, $\mathcal{R}_m$, and $\mathcal{R}_t$, which encode hierarchical, mathematical, and temporal relationships, respectively. The reconstructed dataset $\mathcal{X}^{'}
$ is obtained as:
\begin{equation}
    \mathcal{X}^{'}
 = f_{\text{decom}}\bigl( \mathcal{X}_{\text{c}}^{'}
, \mathcal{R}_h, \mathcal{R}_m, \mathcal{R}_t \bigr),
\end{equation}
where $f_{\text{decom}}: \mathcal{X}_{\text{c}}^{'}
 \times \mathcal{R}_h \times \mathcal{R}_m \times \mathcal{R}_t \rightarrow \mathcal{X}^{'}
$ is the decompression function. Here, $\mathcal{X}^{'}
$ represents the dataset that faithfully reconstructs the original data structure, while preserving the hierarchical, mathematical, and temporal relationships encoded by $\mathcal{R}_h$, $\mathcal{R}_m$, and $\mathcal{R}_t$.

\section{Experimental Settings}
\label{sec:setting}

This section outlines our comprehensive experimental framework for evaluating two key aspects: (1) LLM-based reasoning of inter-column relationships and (2) synthetic tabular data generation. We detail our methodology using real-world industrial datasets, benchmark comparisons against state-of-the-art generation techniques, and complete implementation specifications including hyperparameter configurations.

\section{Experimental Settings}

\subsection{Datasets}
\label{sec:dataset}

\begin{table}[th]
\centering
\caption{Statistics of the two evaluation datasets for generation tasks.} 
    \label{tab:exp-dataset}
\resizebox{0.7\textwidth}{!}{%
\begin{tabular}{rccc|cc}
\toprule
Dataset &
  \begin{tabular}[c]{@{}c@{}}\# of \\ Rows\end{tabular} &
  \begin{tabular}[c]{@{}c@{}}\# of \\ Num. Feat.\end{tabular} &
  \begin{tabular}[c]{@{}c@{}}\# of \\ Cat. Feat.\end{tabular} &
  \begin{tabular}[c]{@{}c@{}}Size of \\ Training Set\end{tabular} &
  \begin{tabular}[c]{@{}c@{}}Size of \\ Test Set\end{tabular} \\ \hline\hline
Retail &
  172,765 &
  26 &
  15 &
  155,488 &
  17,277 \\
Purchasing &
  29,590 &
  14 &
  7 &
  26,631 &
  2,959 \\ \bottomrule
\end{tabular}%
}
\end{table}

Our evaluation framework assesses both logical reasoning generalizability and synthetic data generation quality. 
For reasoning tasks, we first fine-tune/prompt-design our model using two open-source datasets: Retail dataset (from \textit{DataCo})~\citep{dataco} and MIMIC-III's ICU stays data~\citep{mimic}. We then evaluate on three test sets: (1) a private Purchasing dataset for procurement scenarios, (2) the UCI Adult dataset \citep{adult}, and (3) the rest of whole MIMIC-III subsets\citep{mimic} to validate the robustness of LLM-TabLogic. For synthetic data generation evaluation, we focus those two key supply chain datasets: the Retail dataset and the Purchasing dataset. 
The statistics of these two industrial datasets are presented in \autoref{tab:exp-dataset}.

We utilize the Retail dataset (from \textit{DataCo})~\citep{dataco} and MIMIC-III's ICU stays data~\citep{mimic} for our analysis. 
Evaluation is performed on three test sets: (1) a proprietary Purchasing dataset for procurement scenarios, (2) the UCI Adult dataset~\citep{adult}, and (3) additional subsets of MIMIC-III. 
For synthetic data generation, we focus on two key supply chain datasets: the Retail dataset~\citep{dataco} and the Purchasing dataset. 
Statistics for these industrial datasets are summarized in \autoref{tab:exp-dataset}.

As shown in \autoref{tab:exp-dataset}, the Retail dataset, with over 150,000 samples, captures the complex interactions between suppliers, customers, and logistics, integrating geographical, temporal, financial, and product-related information. It includes 26 numerical and 15 categorical columns, modeling transactional activities and the relationships between variables such as payment type, profit per order, and sales per customer. 
The dataset’s complexity arises from the interplay between elements like product categories, customer segments, and order details interact across various levels, influencing sales, pricing, and profitability. 
Additionally, geographic data, such as customer and order locations, links regional sales patterns to product performance.
Similarly, the Purchasing dataset reflects the complex dynamics of procurement, with temporal, categorical, and numerical features. 
It models order cycles, lead times, and stakeholder interactions, with financial attributes like prices, currencies, and discounts adding additional layer of complexity by capturing transactional values and tax implications. 
This combination of diverse variables and complex inter-column relationships makes both datasets and {\it particularly suitable} for evaluating how well generative models preserve inter-column relationships in real-world tabular data.
Moreover, these datasets presents a significant challenge for existing tabular data generation methods, which are primarily designed to enhance machine learning efficiency rather capture inter-column relationships.

\subsection{Baselines}\label{sec:baselines}

We compare our proposed method, LLM-TabLogic, against five representative approaches, including:
\begin{itemize}[nosep]
    \item {\tt SMOTE} \citep{chawla2002smote}, an interpolation-based method originally introduced in the early 2000s to address dataset imbalance, which continues to outperform many modern generative models \citep{margeloiu2024tabebm}.
    
    \item {\tt CTGAN} \citep{xu2019modeling}, a widely used GAN-based method known for its strong performance and industrial relevance.

    \item {\tt TabDDPM} \citep{kotelnikov2023tabddpm} and {\tt TabSyn} \citep{zhang2023mixed}, two diffusion model-based approaches leveraging iterative noise refinement for enhanced data synthesis---TabDDPM being the first of its kind and TabSyn representing the current state-of-the-art in this category.

    \item {\tt GReaT} \citep{borisov2022language}, an LLM-based method that reformulates tabular data as textual sequences to improve representation learning.
\end{itemize}

These methods span multiple generative paradigms, providing a diverse benchmark to evaluate how LLM-TabLogic performs relative to state-of-the-art techniques across different modeling approaches.

\subsection{Synthetic Data Evaluation Framework} 
\begin{figure*}[t]
    \centering
    \includegraphics[width=\textwidth]{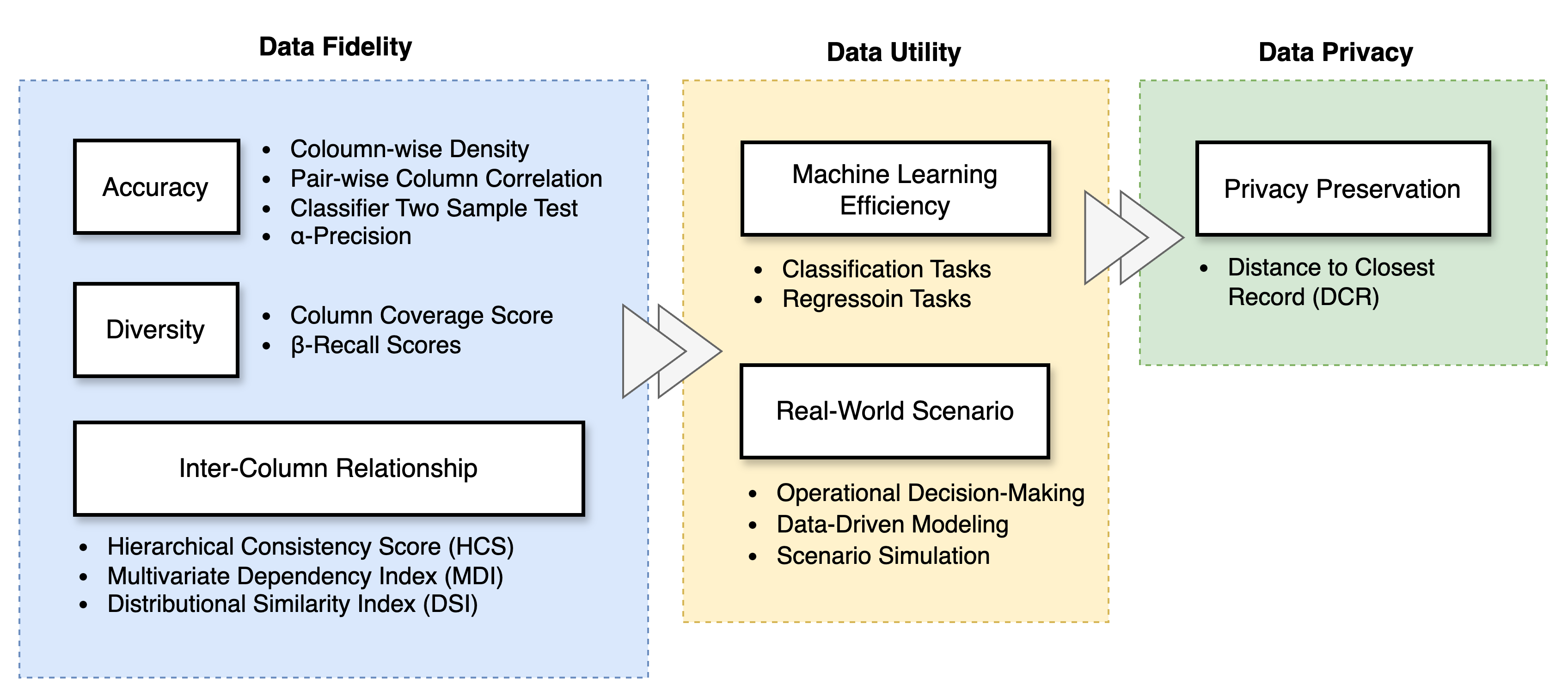}
    \caption{Real-world synthetic tabular data evaluation framework.}
    \label{fig:evaluation_framework}
\end{figure*}

We introduce an evaluation framework to assess the performance of tabular data generation methods.
As illustrated in \autoref{fig:evaluation_framework}, the framework consists of three key assessment dimensions: 
data {\it fidelity}, 
data {\it utility}, and 
data {\it privacy}, each includes multiple sub-dimensions evaluated through a set of performance metrics.
Data fidelity serves as the foundation, assessing whether the generated data accurately preserves the statistical properties, diversity, and inter-column relationships of real data. 
This is essential for ensuring the reliability and practical applicability of synthetic data.
Building on fidelity, data utility examines the effectiveness of synthetic data in downstream tasks, including both machine learning tasks and broader real-world use cases.
This dimension ensures the generated data is effective in real-world scenarios, such as business analytics, industrial simulations, and decision-making.
In addition to data fidelity and utility, privacy is a crucial dimension in evaluating synthetic data generation.
It measures the extent to which sensitive information is retained in the synthetic data, ensuring it does not compromise confidentiality or pose re-identification risks.
Metrics like Distance to Closest Record (DCR) can be adopted to quantify privacy preservation.
Together, these three dimensions would quantify whether synthetic data is statistically accurate, practically useful, and privacy-preserving---the essential aspects that synthetic data must be satisfied in real-world industrial use cases.
We detail these three dimensions in the following sections.

\subsubsection{Data Fidelity}
\label{sec:data_fidelity}
Data fidelity includes three sub-dimensions: 
accuracy, 
diversity, and 
inter-column relationship, as detailed in the following sections.

\paragraph{Accuracy}\label{par:dim_accuracy}
Accuracy evaluates the degree to which synthetic data replicate the characteristics and patterns of real data, considering both low- and high-order statistics.
We use column-wise distribution density estimation and pairwise column correlation for measuring low-order statistics, as described in \citet{zhang2023mixed}. 
Specifically, Kolmogorov-Smirnov Test (KS test) and Total Variation Distance are employed to estimate column-wise density for continuous and categorical columns, respectively.
Pearson correlation and contingency similarity are used to calculate pairwise column correlation for continuous and  categorical columns, respectively.
For accuracy in terms of high-order statistics, we employ 
$\alpha$-Precision \citep{alaa2022faithful}, which quantifies the local similarity between synthetic and real data points.
This metric ensures that the synthetic data accurately represents the feature space of the real data, preserving fine-grained statistical properties, which is particularly crucial for industrial applications.  
Additionally, we use the Classifier Two Sample Test (C2ST) \citep{zein2022tabular} to the similarity between synthetic and real data. 
Higher C2ST scores indicate a greater similarity to real data, making it more realistic to a classifier.

\paragraph{Diversity}\label{par:dim_diversity}
Beyond accurately capturing data distributions, diversity plays a crucial role in evaluating synthetic data quality.
This is particularly vital in industrial applications, where synthetic data must preserve all relevant variations and categories from the real dataset without omitting critical aspects. 
Both low- and high-order statistics are employed for evaluating diversity.
It is crucial to ensure that all possible categories and value ranges from the real data are sufficiently represented in the synthetic data.
To quantify this, we adopt average coverage score, a low-order statistics that evaluates how comprehensively a given region or feature space is covered by the synthetic data. 
It is mathematically defined as:
\begin{equation}
    C_j = 
    \begin{cases}
    \frac{1}{N} \sum\limits_{j=1}^{N} \mathbb{1}\left[ \min(X_i) \leq X^{'}_i(j) \leq \max(X_i) \right] & \text{if } X^{'}_i\text{ is continuous}, \\
    \frac{1}{N} \sum\limits_{j=1}^{N} \mathbb{1}\left[ X^{'}_i(j) \in X_i \right] & \text{if } X^{'}_i \text{ is categorical}.
    \end{cases}
\label{eq:coverage_score}
\end{equation}
where \( X_i \) represents the values of feature \( i \) in the real dataset, \( X^{'}_i(j) \) represents the \( j \)-th row for feature \( i \), \( \mathbb{1}{\left[ \cdot \right]} \) is the indicator function that returns 1 if the condition inside the brackets is true and 0 otherwise, \( N \) is the total number of synthetic data points.

We also incorporate \(\beta\)-Recall \citep{alaa2022faithful}, a high-order statistics, to evaluate diversity.
This metric ensures that edge cases and less common combinations of features are adequately represented in the synthetic data. 
Such coverage is crucial for generating a more robust synthetic dataset, one that captures the full spectrum of variations present in real-world data, including uncommon and edge-case scenarios that might otherwise be overlooked.

\paragraph{Inter-Column Relationship}
\label{par:dim_inter_col_relationship}
Finally, we evaluate inter-column logical relationship preservation \citep{long2025evaluating}, which in turn consists of two key dimensions: consistency and dependency. 
These aspects are critical in determining the quality of synthetic tabular data.
The evaluation is based on three metrics: 
\begin{itemize}[nosep]
    \item {\tt Hierarchical Consistency Score (HCS)}: 
    Measures the preservation of hierarchical relationships, such as city-to-country mappings, by quantifying the proportion of valid attribute groupings.

    \item {\tt Multivariate Dependency Index (MDI)}: 
    Assesses whether synthetic data maintains predefined functional dependencies, ensuring logical consistency across related attributes.

    \item {\tt Distributional Similarity Index (DSI)}: 
    Captures implicit dependencies by comparing the log-likelihoods of Gaussian Mixture Models fitted to real and synthetic datasets, identifying subtle structural deviations.
\end{itemize}

These dimensions collectively ensure that synthetic data retains key structural relationships, making it suitable for downstream applications in industrial contexts.

\subsubsection{Data Utility}
The utility of synthetic data is crucial in real-world industrial applications, where it can be applied to machine learning taskssuch as machine failure prediction and delivery delay prediction, both of which require high levels of accuracy for operational success.
As shown in previous studies, Machine Learning Efficiency is a key factor in evaluating data utility.

To assess the effectiveness of synthetic data in machine learning tasks, we follow an evaluation process widely adopted in prior works \citep{lee2023codi,kotelnikov2023tabddpm,zhang2023mixed}. 
First, generative models are trained on the real dataset to produce synthetic datasets of the same size as the real dataset.
Machine learning models, such as an XGBoost-based classifier or regressor \citep{chen2016xgboost}, are then trained on the {\it synthetic} data but evaluated on {\it real} dataset to measure generalization performance.
For classification tasks, we evaluate performance using metrics like the Area Under the Receiver Operating Characteristic Curve (AUC) and the F1 score. 
For regression tasks, we assess predictive accuracy using the Coefficient of Determination ($R^2$), Root Mean Squared Error (RMSE), and Mean Absolute Error (MAE).
Evaluating data utility is essential as it directly reflects the practical value of synthetic data. 
High test accuracy, when the model is trained on synthetic data, indicates that the synthetic dataset effectively preserves the underlying patterns and structures of the real data. 
This evaluation is crucial for validating the applicability of synthetic data in real-world industrial machine learning tasks.

Moreover, many industrial applications—such as operational decision-making, robust data-driven modeling, and scenario simulations—require synthetic data to be sufficiently realistic in mimicking real data to ensure reliable and actionable insights.
Therefore, evaluating the utility of synthetic data {\it must} go beyond machine efficiency to its applicability in real-world scenarios. 
This involves evaluating the logical relationships within the data to ensure alignment with real-world operational dynamics, such as constructing retail networks, adhering to domain-specific rules, or simulating transportation events.
By measuring how well the synthetic data meets these criteria, we can determine its reliability in driving actionable insights and realistic simulations.

\subsubsection{Data Privacy}
Synthetic data must preserve privacy by preventing the unintended disclosure of sensitive information from real-world industrial datasets and mitigating risks such as data re-identification and membership inference attacks.
Therefore, assessing data privacy is a crucial dimension in evaluating the quality and reliability of synthetic data.
To quantify privacy preservation, we employ the Distance to Closest Record (DCR) metric \citep{liu2024scaling}, which measures the pairwise distance between real and synthetic data points. 
This metric determines whether the synthetic data is generated based on the overall distribution of the real data, rather than replicating specific records from the training dataset. 
The DCR between each synthetic data point \( s_i \in S \) and all real data points \( r_j \in R \) is calculated using the \( L_1 \) (Manhattan) distance, which is the sum of the absolute differences between corresponding features.
For each synthetic data point \( s_i \), the minimum distance to any real data point \( r_j \) is computed and denoted as \( d_{\text{min}}(s_i, R) \), representing the closest real data point to \( s_i \). 
Similarly, the minimum distance between each synthetic data point \( s_i \) and all test data points \( t_k \in T \) is computed in the same manner, with the minimum distance denoted as \( d_{\text{min}}(s_i, T) \), representing the closest test data point to \( s_i \).
The DCR is then defined as the proportion of synthetic data points whose minimum distance to the real data is smaller than the minimum distance to the test data:
\begin{equation}
\label{eq:DCR}
    \text{DCR} = \frac{1}{N} \sum_{i=1}^{N} \mathbb{1}\Bigl( d_{\text{min}}(s_i, R) < d_{\text{min}}(s_i, T) \Bigr)
\end{equation}
where \( N \) is the total number of synthetic data points, and \( \mathbb{1}(\cdot) \) is an indicator function that returns 1 if the condition is satisfied and 0 otherwise.

This DCR score quantifies how closely the synthetic data points resemble the real data, with a {\it higher} score indicating that the synthetic data is closer to real data, which could imply a privacy risk. 
A lower DCR score suggests that the synthetic data points are more {\it distinct} from the real data, thereby better preserving privacy.

\subsection{Experimental Setup} 

\paragraph{\textbf{Data Preparation}}
For the LLM reasoning stage, only the meta data of each tables are input into LLM-TabLogic for logics inference.
For the generation, the two evaluation datasets were first split into training and testing sets, following the separation ratio specified in \autoref{tab:exp-dataset}.
Tabular data generation methods were then trained on the real training set, and the trained models were used to generate synthetic dataset with the same size as the real training set.

\paragraph{\textbf{Evaluation}}
In our experiments, we evaluated the logical reasoning capabilities of both GPT-4 \citep{achiam2023gpt} and DeepSeek-Va R1 \citep{deepseekai2025} models for inferring inter-column relationships, leveraging their distinct strengths to capture complex tabular dependencies. First, we compared GPT-4 \citep{achiam2023gpt} and DeepSeek-Va R1 \citep{deepseekai2025} under deterministic conditions (temperature = 0.1), reporting averaged results from 10 independent responses to measure baseline performance. Second, to assess robustness, we systematically varied the temperature (0.1–0.9) for both models, analyzing how output stability scales with stochasticity. This dual approach not only highlights the comparative performance of the two LLMs but also quantifies their stability under varying levels of stochasticity.
We evaluate LLM hallucination in logical relationship prediction using the F1 score across three benchmark datasets (Retailing, ICUstays from MIMIC-III, and Adult). Our methodology involves: (1) designing and validating prompts on the Retailing and ICUstays datasets, then (2) testing on the Adult, Purchasing and the rest of whole MIMIC-III datasets. The F1 score balances precision ($P=\frac{TP}{TP+FP}$) to quantify hallucination rates (penalizing false positives) and recall ($R=\frac{TP}{TP+FN}$) to assess coverage (penalizing false negatives), providing a unified measure of both prediction accuracy and completeness in inter-column logic inference.

For the data generation tasks, to ensure a robust evaluation, we generated {\it ten} synthetic datasets using each trained tabular generation method. 
And we report mean and standard deviation metrics from 10 sampling runs using the best-validated model.
Each synthetic dataset was then used to train task-specific models---specifically, an XGBoost classifier in our experiments---for downstream tasks such as prediction.

\paragraph{\textbf{Hyperparameters}}
For the Data re

The hyperparameters of the prediction model---XGBoost classifier---were search within the following parameter space: 
number of trees: \{{\tt 100, 200, 300}\}; 
learning rate: \{{\tt 0.1, 0.01, 0.001}\}; 
maximum tree depth: \{{\tt 5, 10}\}; 
minimum child weight: \{{\tt 1, 10}\}; 
and gamma: \{{\tt 0.0, 0.5, 1.0}\}.
Additionally, Grid Search was employed to identify the optimal hyperparameter combination for the XGBoost classifer. For the generative models, we finetune them following their hyperparameter settings from papers

\paragraph{\textbf{Computational Resources}}
We implemented all the experiments in Python. 
The implementations of the tabular data generation methods and the XGBoost classifier were sourced from their original papers. 
For evaluation metrics, we used those available in SDMetrics \footnote{\url{https://docs.sdv.dev/sdmetrics}}; otherwise, we implemented them ourselves.
Our experiments were conducted in the workstation with the following configurations: 
Ubuntu 20.04.6 LTS with the Linux kernel version 5.15.0-113-generic, 
an Intel(R) Xeon(R) Platinum 8368 CPU (2.40 GHz), and 
an NVIDIA GeForce RTX 4090 GPU.
These configurations are well-suited for high-performance computing and machine learning tasks.

\section{Logical Reasoning Results} \label{sec:logic_results}

\begin{figure*}[t]
    \centering
    \begin{subfigure}{0.49\textwidth}
        \centering
        \includegraphics[width=\textwidth]{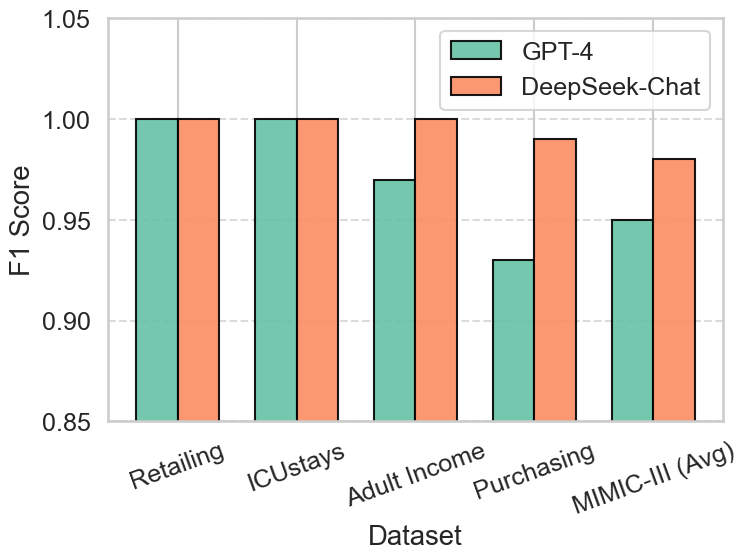}

        \label{fig:model_comparison}
    \end{subfigure}
    \begin{subfigure}{0.49\textwidth}
        \centering
        \includegraphics[width=\textwidth]{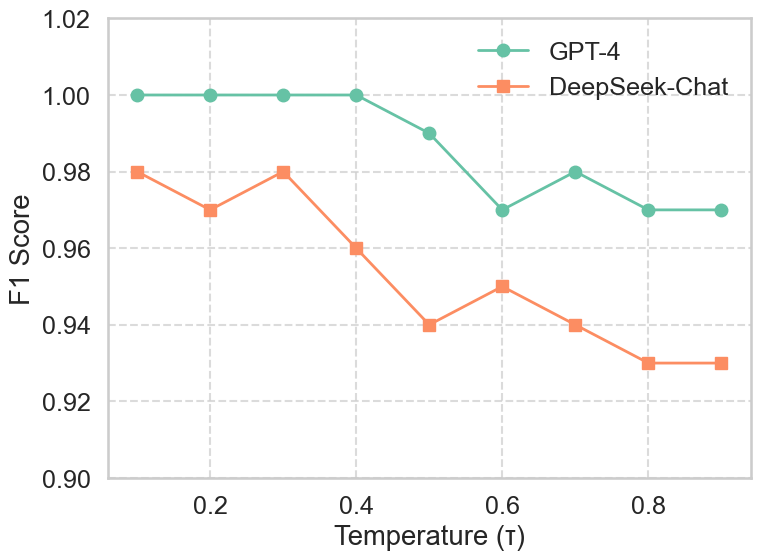}
        \label{fig:temperature_effect}
    \end{subfigure}
    \caption{
        Performance comparison of GPT-4 and DeepSeek-Chat on tabular reasoning tasks. 
        (a) shows F1 scores of GPT-4 and DeepSeek-Chat across industrial datasets. 
        (b) presents the effect of temperature on F1 performance of both models on the Purchasing Datasets.
    }
    \label{fig:tabular_reasoning_results}
\end{figure*}

As shown in Table~\ref{fig:tabular_reasoning_results}, our comparative analysis highlights consistent performance across datasets for both GPT-4 and DeepSeek-Chat. On the Retailing and Transfers datasets, both models achieve perfect F1 scores (1.00), which can be attributed to prompt designs that were specifically fine-tuned for these domains. In contrast, the remaining datasets—Purchasing, MIMIC-III (Avg), and Adult Income—represent unseen data not included during prompt tuning. On these more challenging and generalizable tasks, DeepSeek-Chat outperforms GPT-4 on both the Purchasing dataset (0.98 vs. 0.93) and MIMIC-III average (0.96 vs. 0.91), while both models again achieve perfect scores on Adult Income (1.00). These results underscore DeepSeek-Chat’s robust and adaptable reasoning performance across both fine-tuned and unseen tabular tasks.

\paragraph{\textbf{Temperature Sensitivity Analysis}}
We investigate the impact of temperature variation ($\tau \in [0.1, 1.0]$) on model performance by analyzing F1 score trends. DeepSeek-Chat demonstrates strong robustness, maintaining stable F1 scores across the full temperature range with a mean of $0.79 \pm 0.02$. In contrast, GPT-4 exhibits a notable decline in performance, with its F1 score dropping from 0.82 at low temperatures to 0.71 at higher temperatures. This highlights DeepSeek-Chat's greater stability in reasoning tasks under diverse sampling conditions, whereas GPT-4's performance is more sensitive to decoding randomness.

\section{Data Genertaoin Results} \label{sec:results}

\begin{figure*}[t]
    \centering
    \begin{subfigure}{0.49\textwidth}
        \centering
        \includegraphics[width=\textwidth]{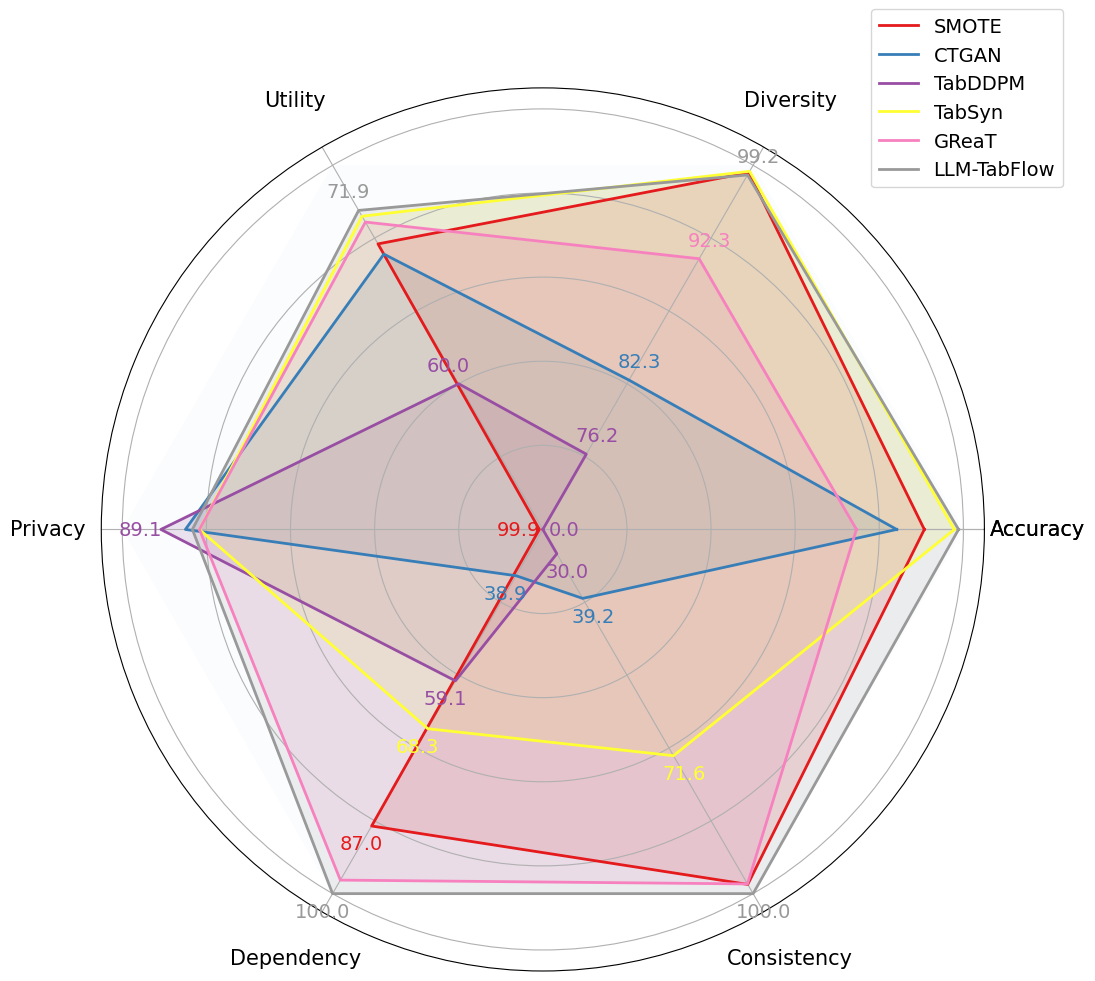}
        \caption{Retailing}
        \label{fig:retail}
    \end{subfigure}
    \begin{subfigure}{0.49\textwidth}
        \centering
        \includegraphics[width=\textwidth]{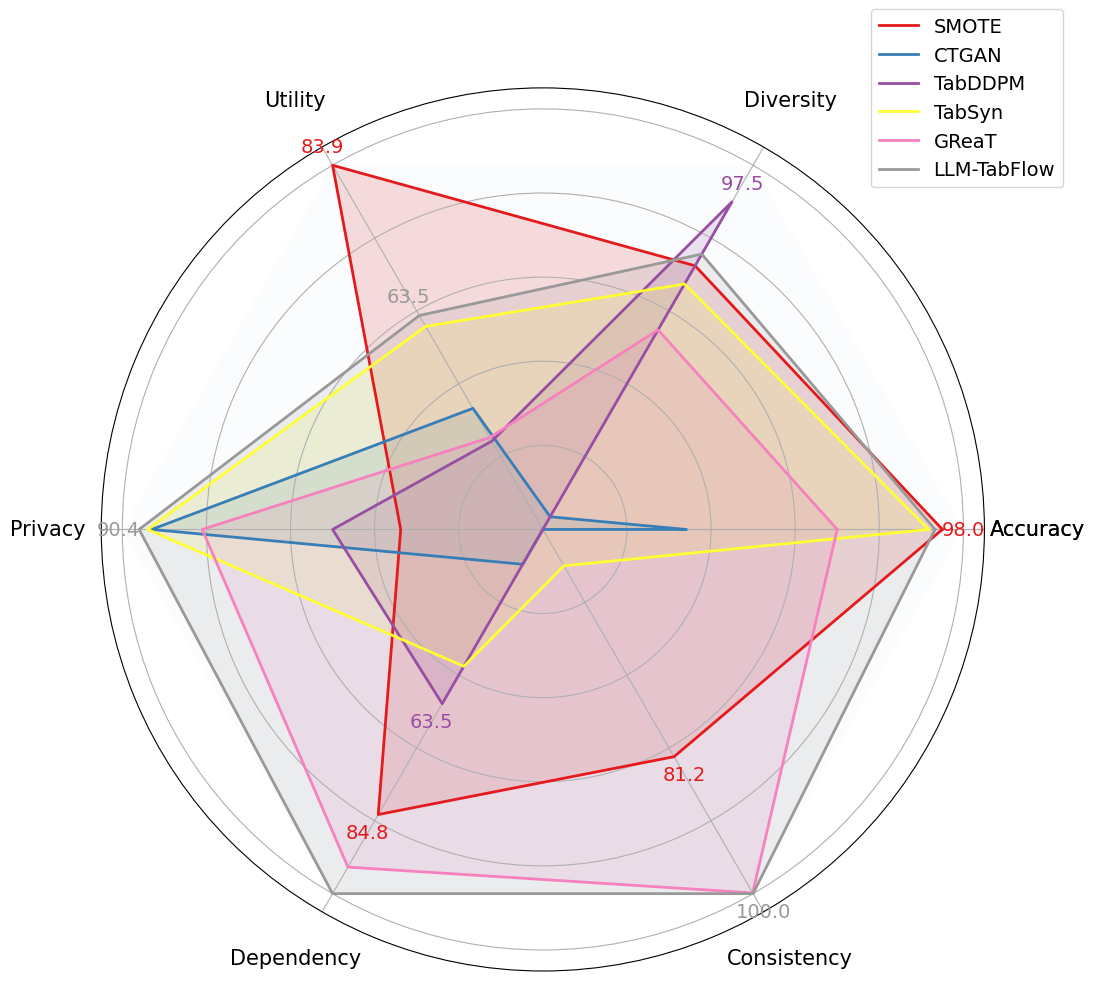}
        \caption{Purchasing}
        \label{fig:purchasing}
    \end{subfigure}
    \caption{
        Radar Charts illustrating the performance of all methods on the two datasets across the six dimensions: Accuracy, Diversity, Utility, Privacy, Consistency, and Dependency. 
        Values closer to the outermost circle indicate better performance. 
        LLM-TabLogic achieves the overall best performance, with the most balanced results across all metrics.
    }
    \label{fig:radar_charts}
\end{figure*}

This section presents the evaluation results for synthetic data produced by our proposed method---LLM-TabLogic---and baseline methods. 
Each evaluation is conducted {\it ten} times, repeating the process of generating and evaluating synthetic data.
The results are reported as the results presented as the mean and standard deviation in the format ${\tt mean} \pm {\tt std dev}$.

\subsection{Overall Evaluation}
\label{sec:over_evaluation}
The overall performance of LLM-TabLogic and the five baseline methods on the two datasets is visualized as Radar charts, as illustrated in \autoref{fig:radar_charts}. 
To provide a more comprehensive evaluation and clearer visualization, the inter-column relationship dimension is divided into two distinct aspects: Consistency and Dependency.
As a result, the radar charts consist of six dimensions: Accuracy, Diversity, Utility, Privacy, Consistency, and Dependency.
Surprisingly, as shown in \autoref{fig:radar_charts}, SMOTE, originally proposed in early 2000s, still outperforms advanced deep generative models in several aspects such as accuracy, diversity, and inter-column relationships, while also achieving strong performance in utility on the Purchasing dataset. 
However, compared to deep generative models such as latent-space or PLM-based methods, this interpolation-based approach heavily relies on real input data for data sampling and often compromising data privacy. 
These limitations significantly restrict its applicability in industrial scenarios involving sensitive data.
In contrast, advanced models such as TabDDPM and CTGAN perform slightly worse than SMOTE across almost all metrics. 
While TabSyn, a state-of-the-art method, surpasses SMOTE in privacy and utility, it struggles to fully capture consistency and dependency in the real data. 
Overall, models relying on latent representation exhibit significant limitations in preserving logical relationships.
Notably, GReaT, a PLM-based model, excels in maintaining strong logical relationships but falls short in other critical dimensions, including data fidelity and utility. 
In contrast, our proposed approach---LLM-TabLogic---achieves {\it near-perfect} results across all six dimensions, ensuring consistency and dependency while excelling in fidelity, utility, and privacy. 
This demonstrates the robustness and versatility of LLM-TabLogic in generating high-quality realistic synthetic tabular data.
We further examine the experimental results in each of the six dimensions in the following sections.

\subsection{Accuracy and Diversity}
\label{sec:acc_div}

\renewcommand{\arraystretch}{1.5} 
\begin{table}[h]
\centering
\caption{ 
Evaluation results of the six methods in terms of accuracy (density estimation, Pearson correlation, and $\alpha$-Precision) and diversity (average coverage score and $\beta$-Recall). 
Higher values of these metrics indicate better performance. 
The best and second-best results are highlighted in {\color{blue}\bf blue} and {\bf black boldfaced}, respectively.
}
\label{tab:accuracy}
\resizebox{\textwidth}{!}{%
\begin{tabular}{cccccccc}
\toprule[1pt]
\multirow{2}{*}{\textbf{Datasets}} &
\multirow{2}{*}{\textbf{Metrics}} &
  \multicolumn{1}{c}{\textbf{Interpolation-based}} &
  \multicolumn{3}{c}{\textbf{Latent Space-based}} &
  \textbf{PLM-based} & \textbf{Ours}\\ \cmidrule(lr){3-3} \cmidrule(lr){4-6} \cmidrule(lr){7-7} \cmidrule(lr){8-8} 
 &  &
  \textbf{SMOTE} &
  \textbf{CTGAN} &
  \textbf{TabDDPM} &
  \textbf{TabSyn} &
  \textbf{GReaT} &
  \textbf{LLM-TabFlow} \\ \hline

\multirow{5}{*}{\textbf{Retailing}} &
Density Estimation & 
  \textcolor{blue}{\textbf{98.02±0.22}} & 90.38±0.23 & 33.11±0.12 & 96.38±0.24 & 89.58±0.32 & \bf{96.46±0.23} \\ 
&
Pairwise Correlation & 
  \textcolor{blue}{\textbf{96.21±0.62}} & 74.41±0.24 & 36.78±0.31 & \textbf{94.81±0.44} & 71.00±1.36 & 92.52±0.42 \\ 

&
$\alpha$-Precision & 
  93.54±0.11 & 88.90±0.24 & 0.00±0.00 & \textbf{98.48±0.21} & 82.18±0.19 & \textcolor{blue}{\textbf{99.15±0.17}} \\

\cmidrule(lr){2-8}
  
&
Average Coverage Score & 
  \textbf{99.41±0.18} & 82.27±0.34 & 76.23±0.23 & \textcolor{blue}{\bf{99.52±0.31}} & 92.34±0.45 & 99.18±0.24 \\ 

&
$\beta$-Recall & 
  \textcolor{blue}{\textbf{72.21±0.13}} & 1.71±0.21 & 0.00±0.00 & 22.62±0.32 & 24.05±0.21 & \textbf{27.85±0.23}\\
\midrule\midrule
\multirow{5}{*}{\textbf{Purchasing}} &
Density Estimation & 
 \textcolor{blue}{\textbf{99.11±0.15}} & 86.36±0.34 & 48.01±0.44 & \bf{98.61±0.35} & 92.72±0.38 & 98.14±0.47 \\ 
&
Pairwise Correlation & 
  \textcolor{blue}{\textbf{98.83±0.21}} & 82.25±0.33 & 51.00±0.34 & \textbf{98.63±0.20} & 74.18±3.22 & \textbf{\textit{92.34±1.12}} \\
&

$\alpha$-Precision & 
  \textcolor{blue}{\textbf{98.01±0.32}} & 73.67±0.22 & 4.16±0.31 & 96.73±0.37 & 88.00±0.36 & \textbf{97.29±0.25} \\ 
\cmidrule(lr){2-8}
  
&
Average Coverage Score & 
  \textbf{93.19±2.34} & 75.87±0.13 & 97.49±0.80 & 91.85±0.45 & 88.72±2.97 & \textcolor{blue}{\textbf{93.92±0.23}} \\

&
$\beta$-Recall & 
 \textcolor{blue}{\textbf{62.79±0.63}} & 4.23±0.38 & 0.03±0.04 & 15.30±0.37 & \bf{36.16±0.32} & 17.32±0.21 \\
\bottomrule

\end{tabular}%
}
\end{table}

\begin{figure*}[th]
    \centering
    \begin{subfigure}{0.32\textwidth}
        \centering
        \includegraphics[width=\textwidth]{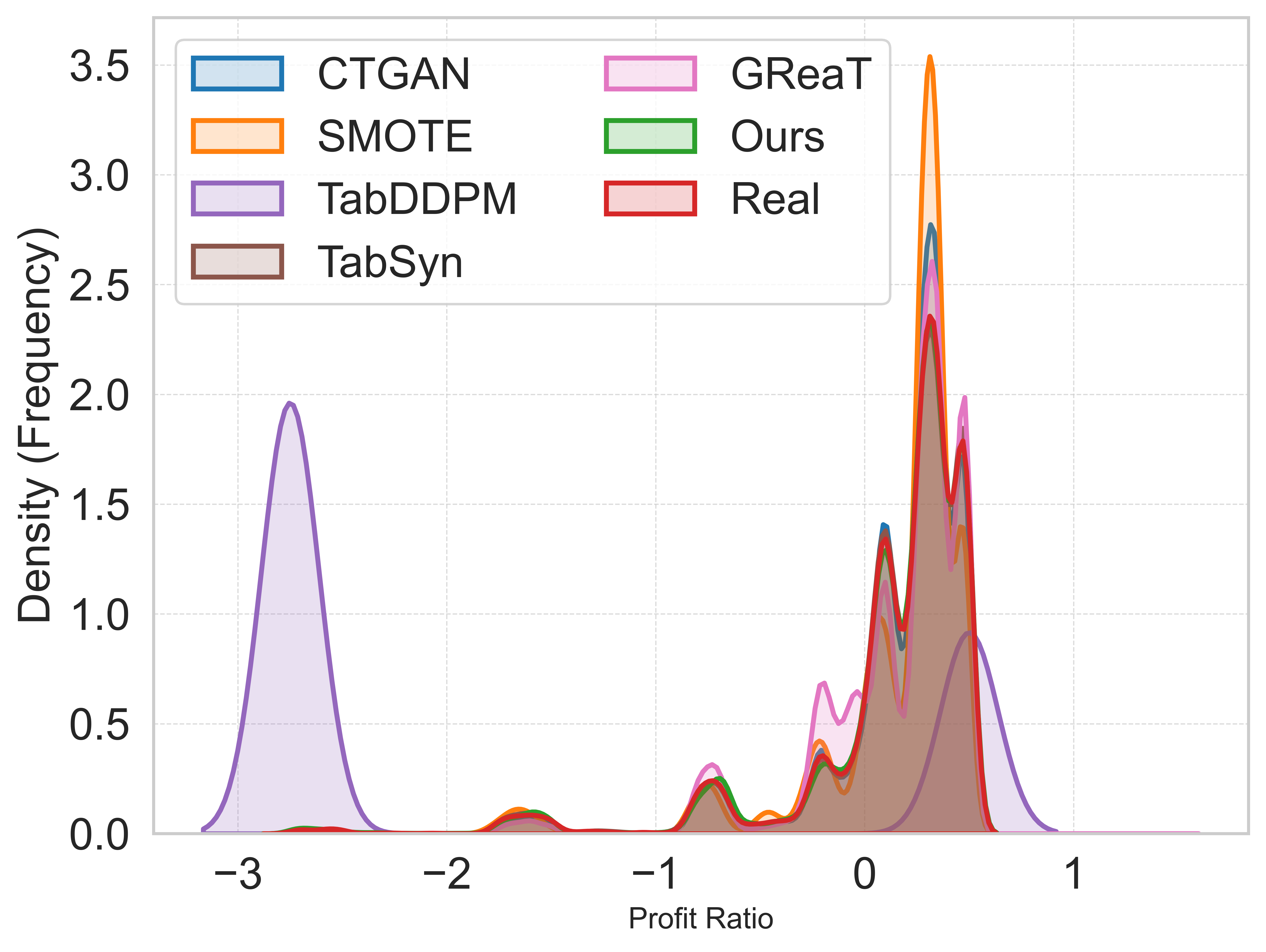}
        \caption{Item Profit Ratio}
    \end{subfigure} 
    \begin{subfigure}{0.33\textwidth}
        \centering
        \includegraphics[width=\textwidth]{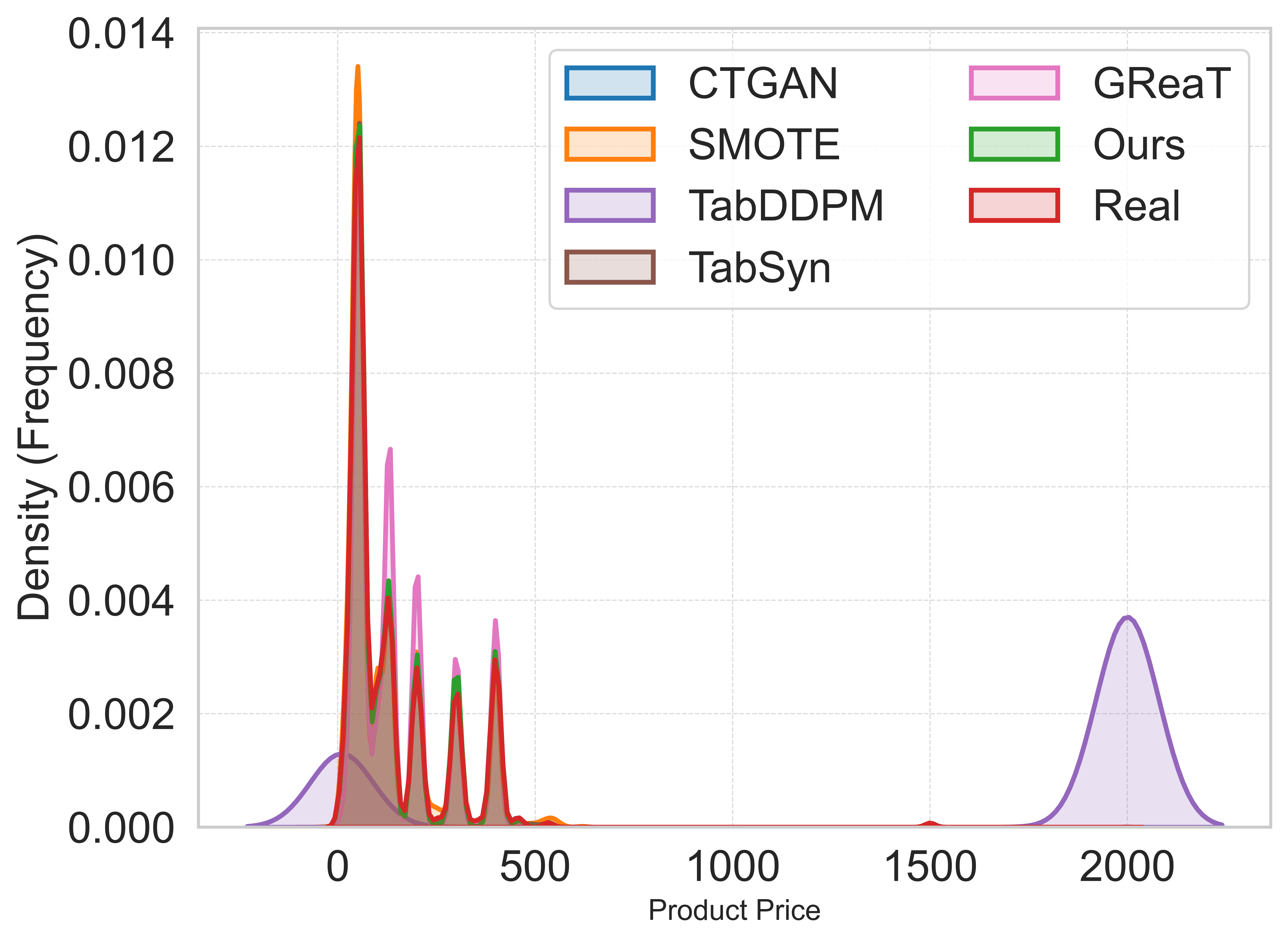}
        \caption{Product Price}
    \end{subfigure} 
    \begin{subfigure}{0.325\textwidth}
        \centering
        \includegraphics[width=\textwidth]{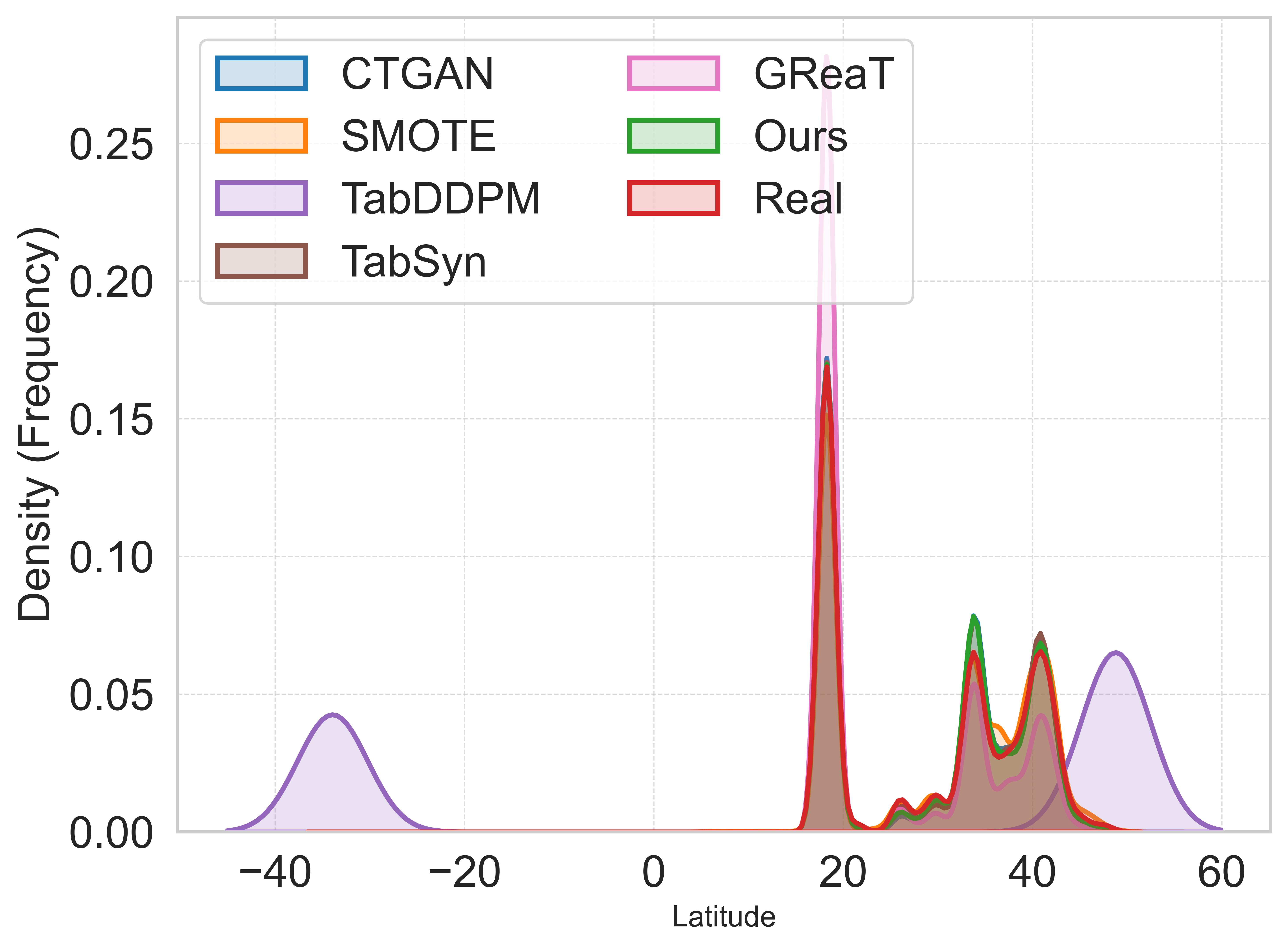}
        \caption{Latitude}
    \end{subfigure}    
    \caption{Density plots for the three {\bf continuous} columns ({\tt item profit ratio}, {\tt product price}, and {\tt latitude}), comparing the distribution of real data and their synthetic counterparts generated by different methods.
    Curves that more closely align with the real data indicate better performance. 
    Both LLM-TabLogic and TabSyn exhibit distributions that closely match the real data, outperforming other methods.}
    \label{fig:accuracy1}
\end{figure*}

\begin{figure*}[ht!]
    \centering
    \begin{subfigure}[t]{0.32\textwidth}
        \centering
        \includegraphics[width=\textwidth, height=5cm]{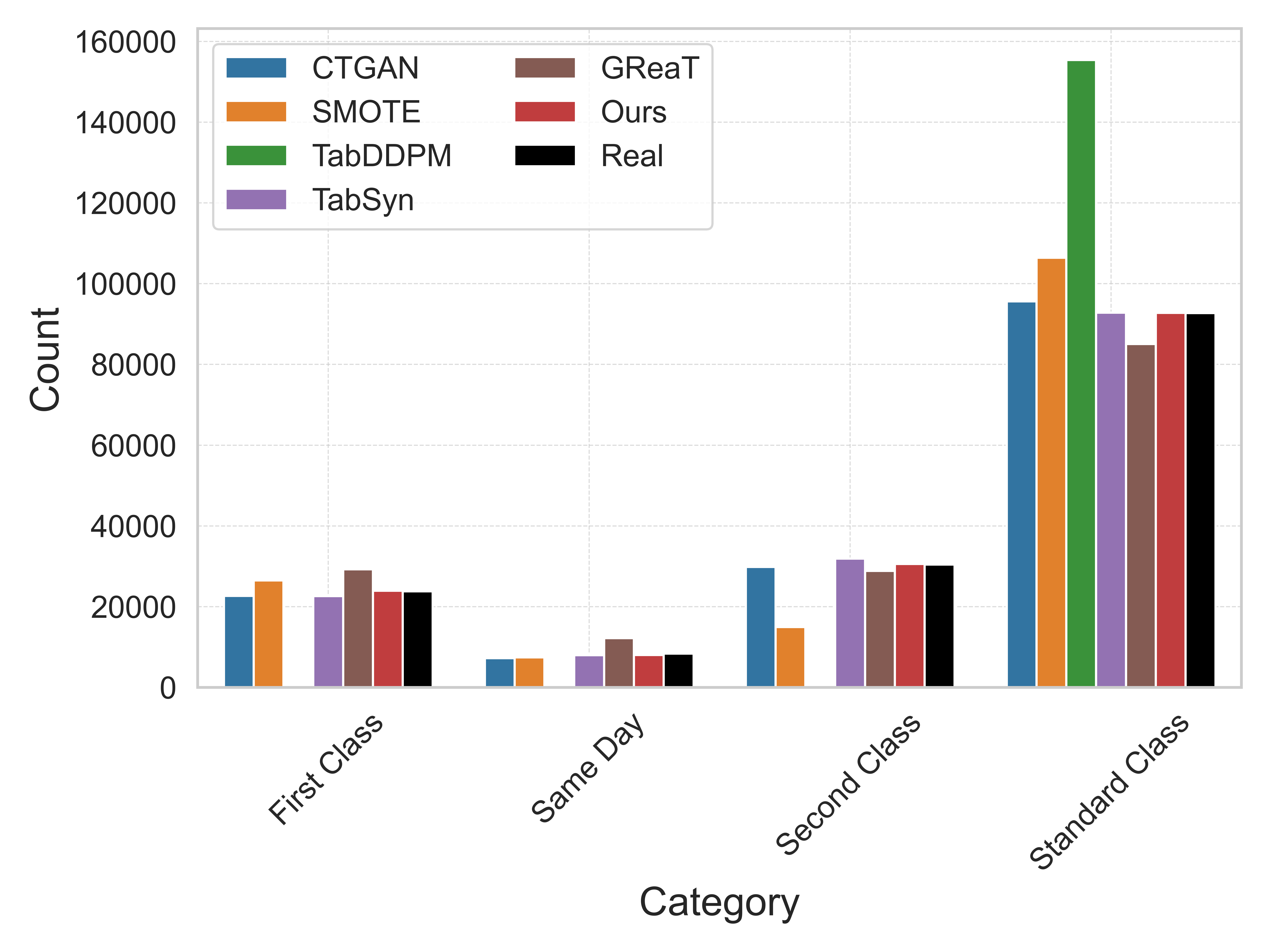} 
        \caption{Shipping Mode}
    \end{subfigure} 
    \begin{subfigure}[t]{0.32\textwidth}
        \centering
        \includegraphics[width=\textwidth, height=5cm]{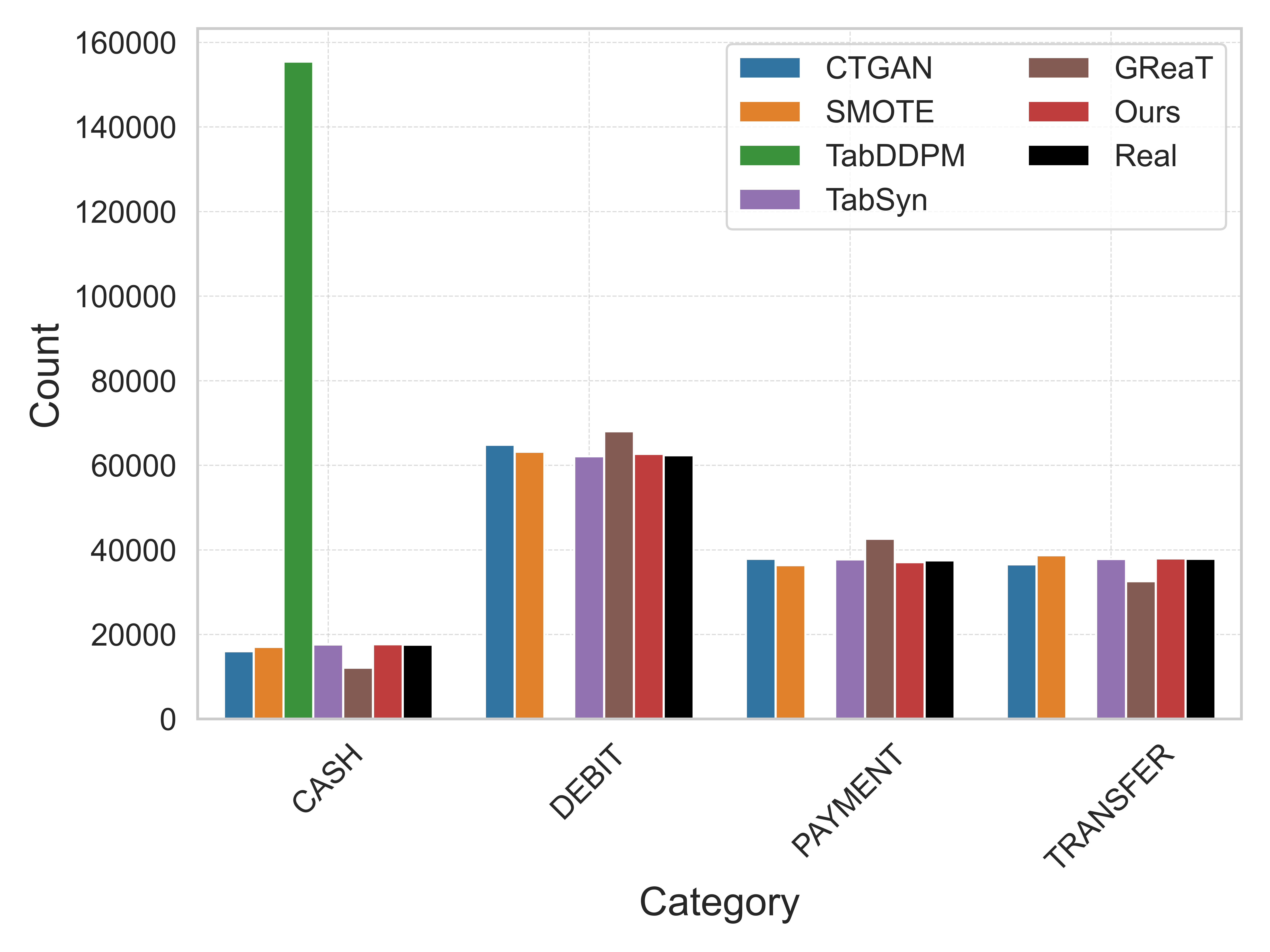} 
        \caption{Payment Type}
    \end{subfigure} 
    \begin{subfigure}[t]{0.32\textwidth}
        \centering
        \includegraphics[width=\textwidth, height=5cm]{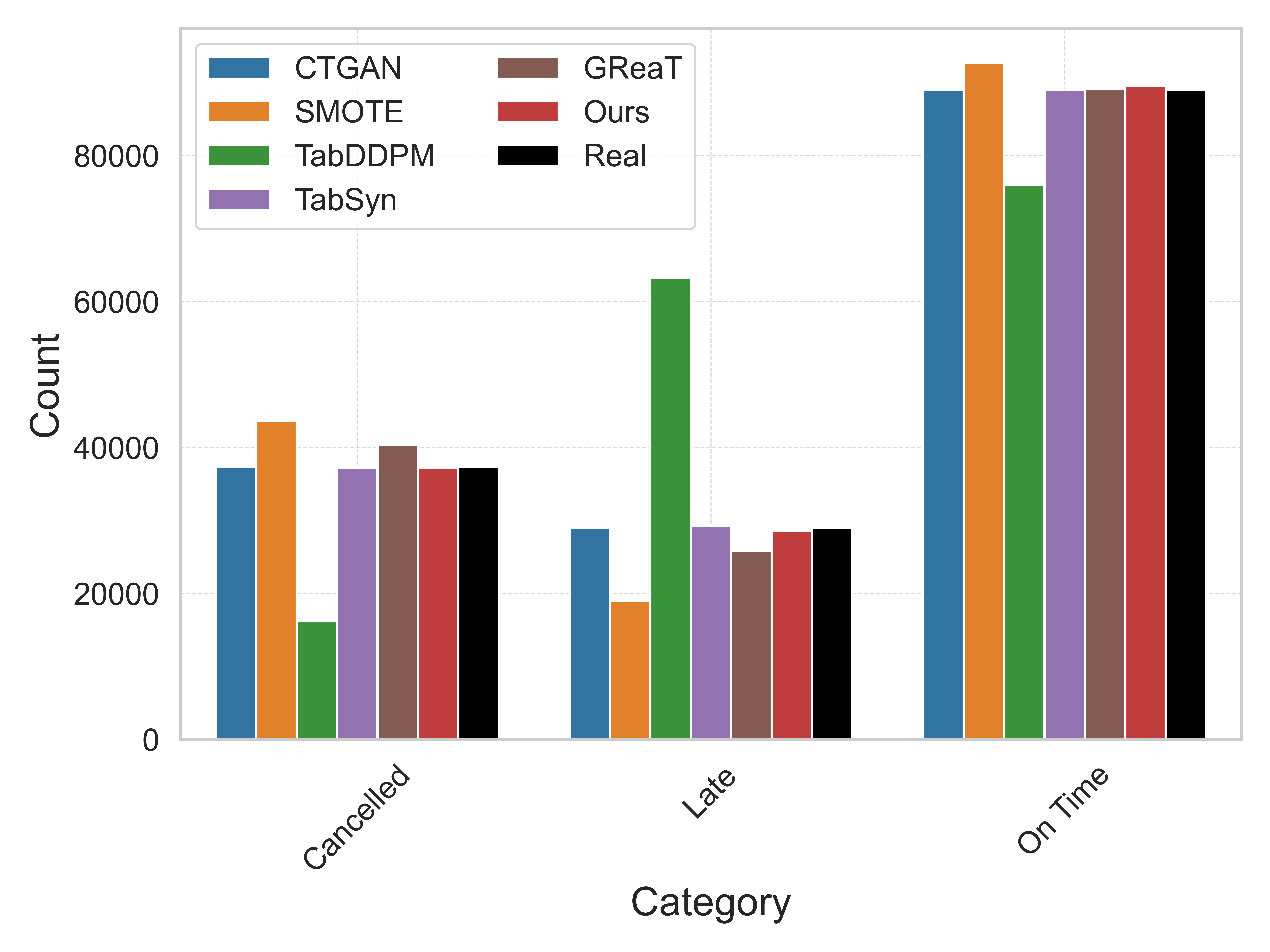} 
        \caption{Order Status}
    \end{subfigure}
    \caption{
        Distribution plots for the three {\bf categorical} columns ({\tt shipping mode}, {\tt payment type}, and {\tt order status}), comparing synthetic data to real data. 
        Distributions that closely match the real data indicate superior performance. 
        Both LLM-TabLogic and TabSyn exhibit distributions that are significantly closer to the real data compared to other methods.
    }
    \label{fig:accuracy2}
\end{figure*}

\begin{figure*}[h!]
    \centering
    \begin{subfigure}[t]{0.32\textwidth}
        \centering
        \includegraphics[width=\textwidth]{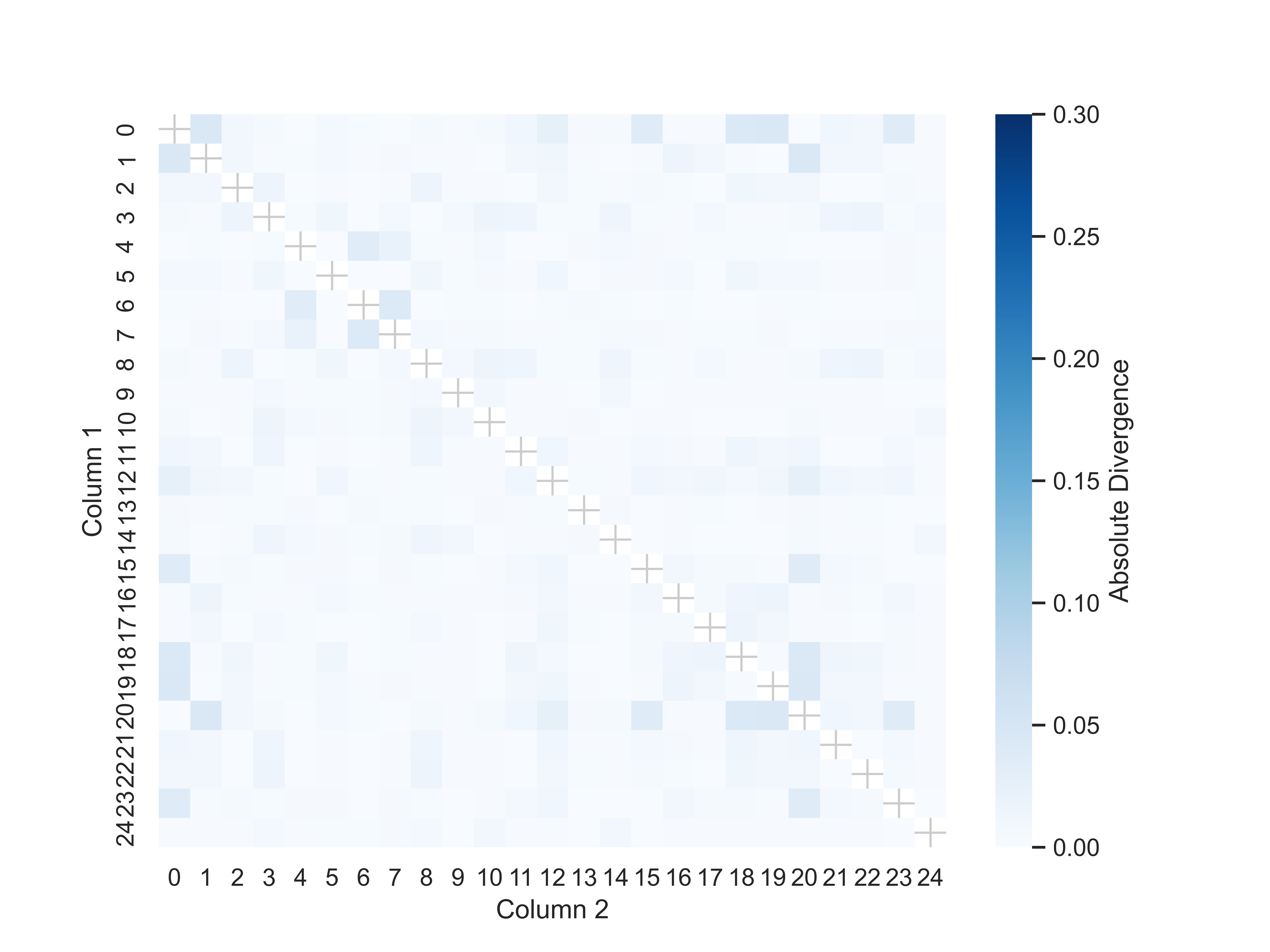}
        \caption{SMOTE}
    \end{subfigure} 
    \begin{subfigure}[t]{0.32\textwidth}
        \centering
        \includegraphics[width=\textwidth]{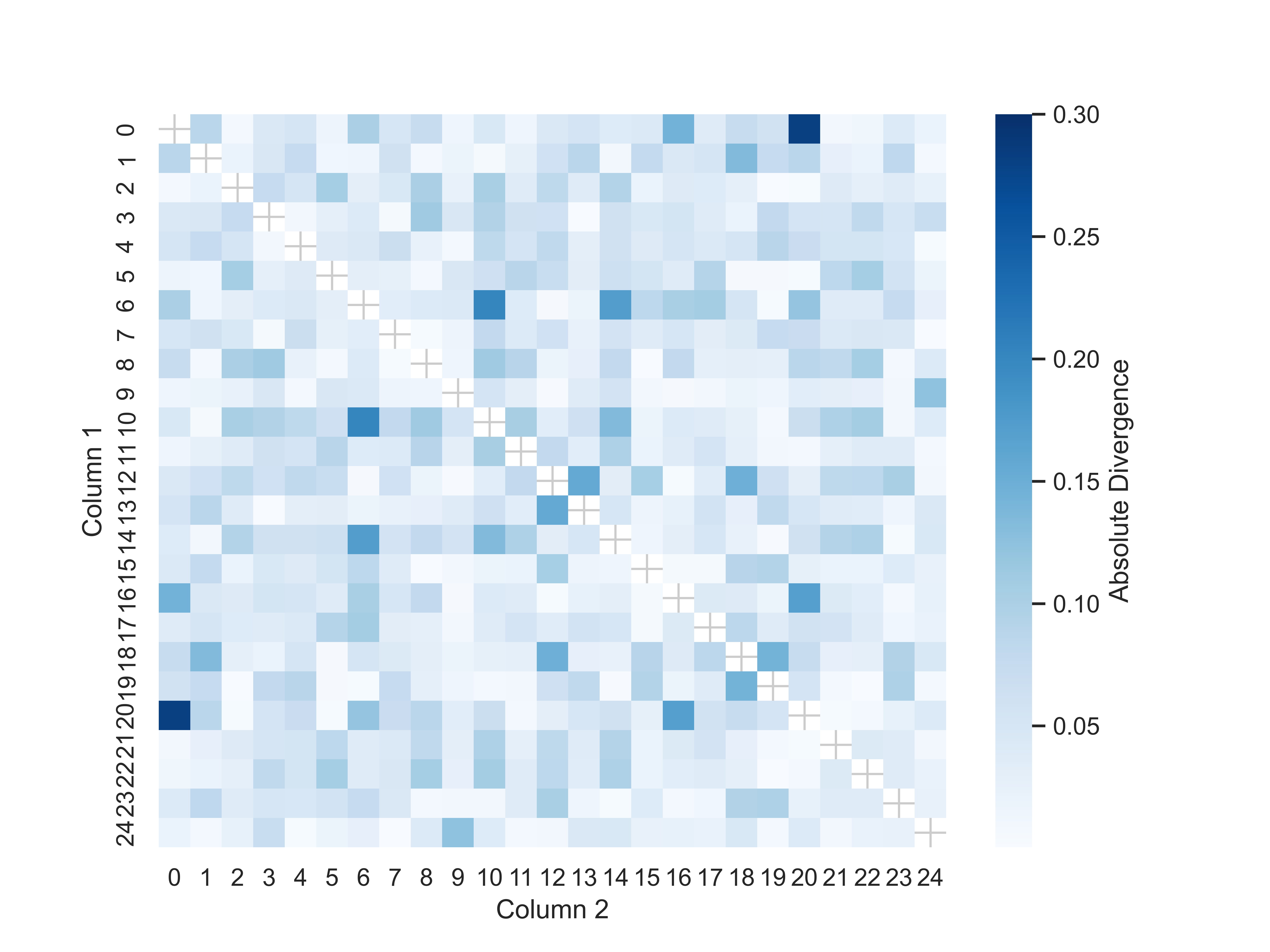}
        \caption{CTGAN}
    \end{subfigure} 
    \begin{subfigure}[t]{0.32\textwidth}
        \centering
        \includegraphics[width=\textwidth]{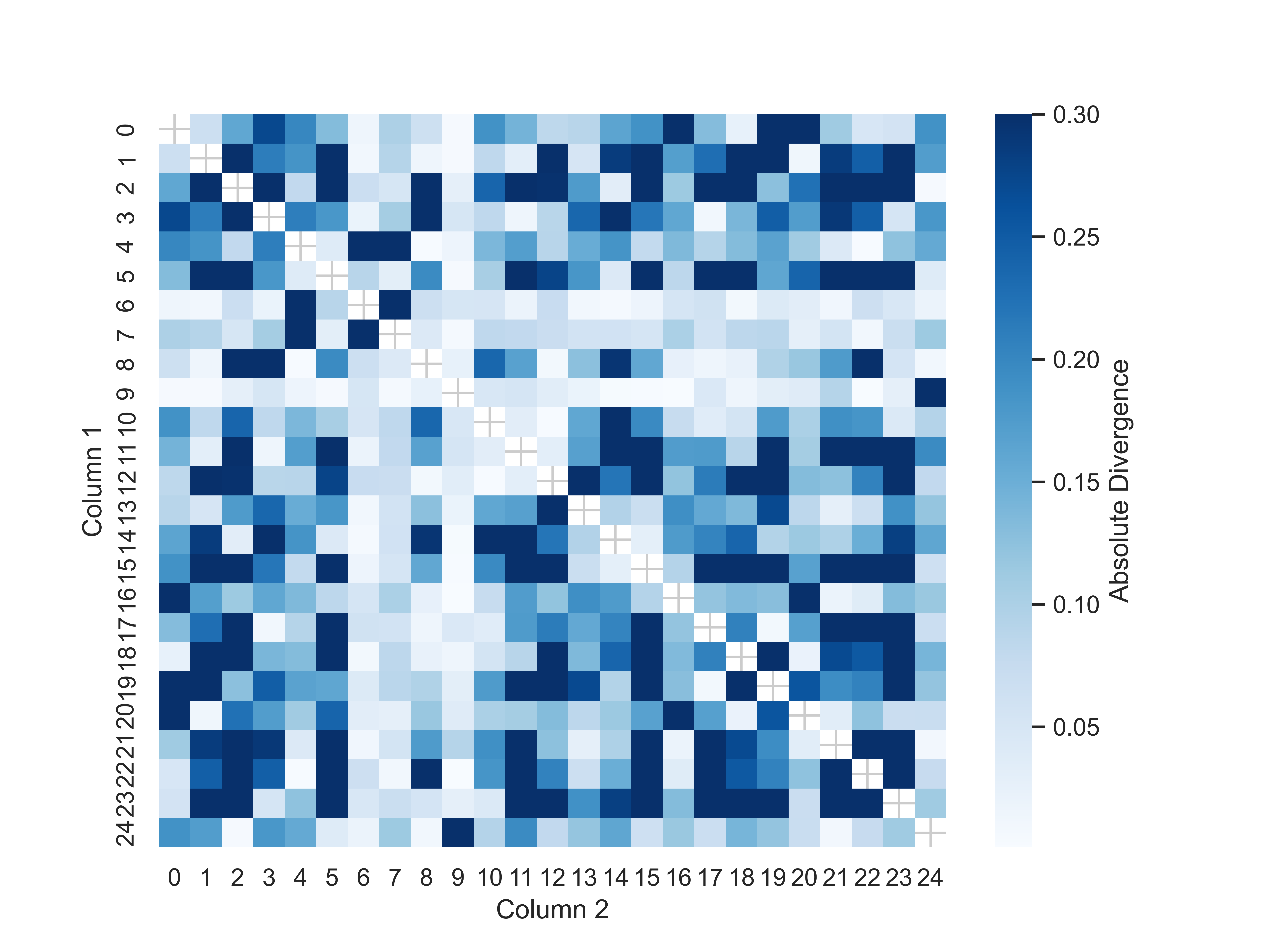}
        \caption{TabDDPM}
    \end{subfigure} 
    \begin{subfigure}[t]{0.32\textwidth}
        \centering
        \includegraphics[width=\textwidth]{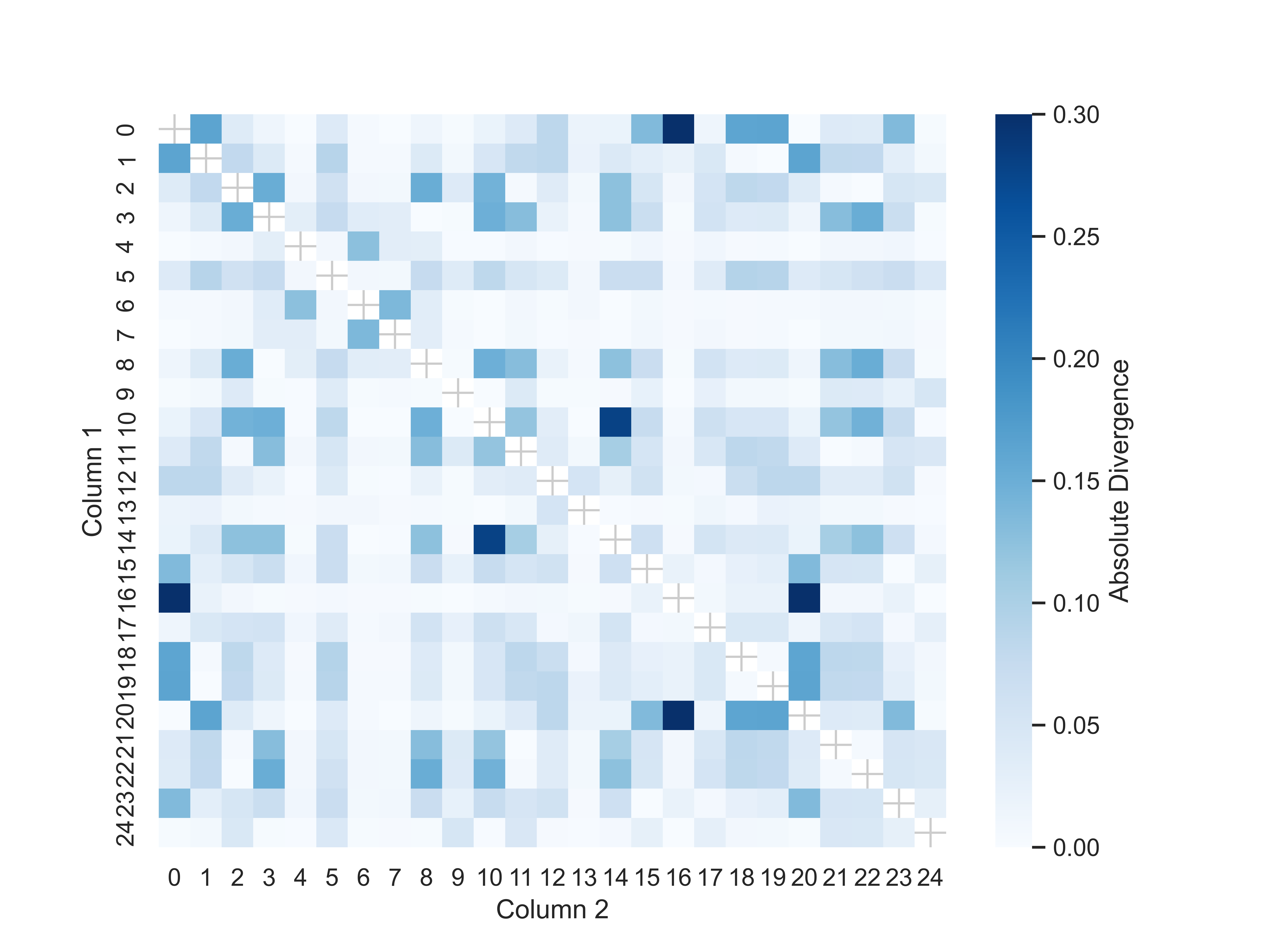}
        \caption{GReaT}
    \end{subfigure} 
    \begin{subfigure}[t]{0.32\textwidth}
        \centering
        \includegraphics[width=\textwidth]{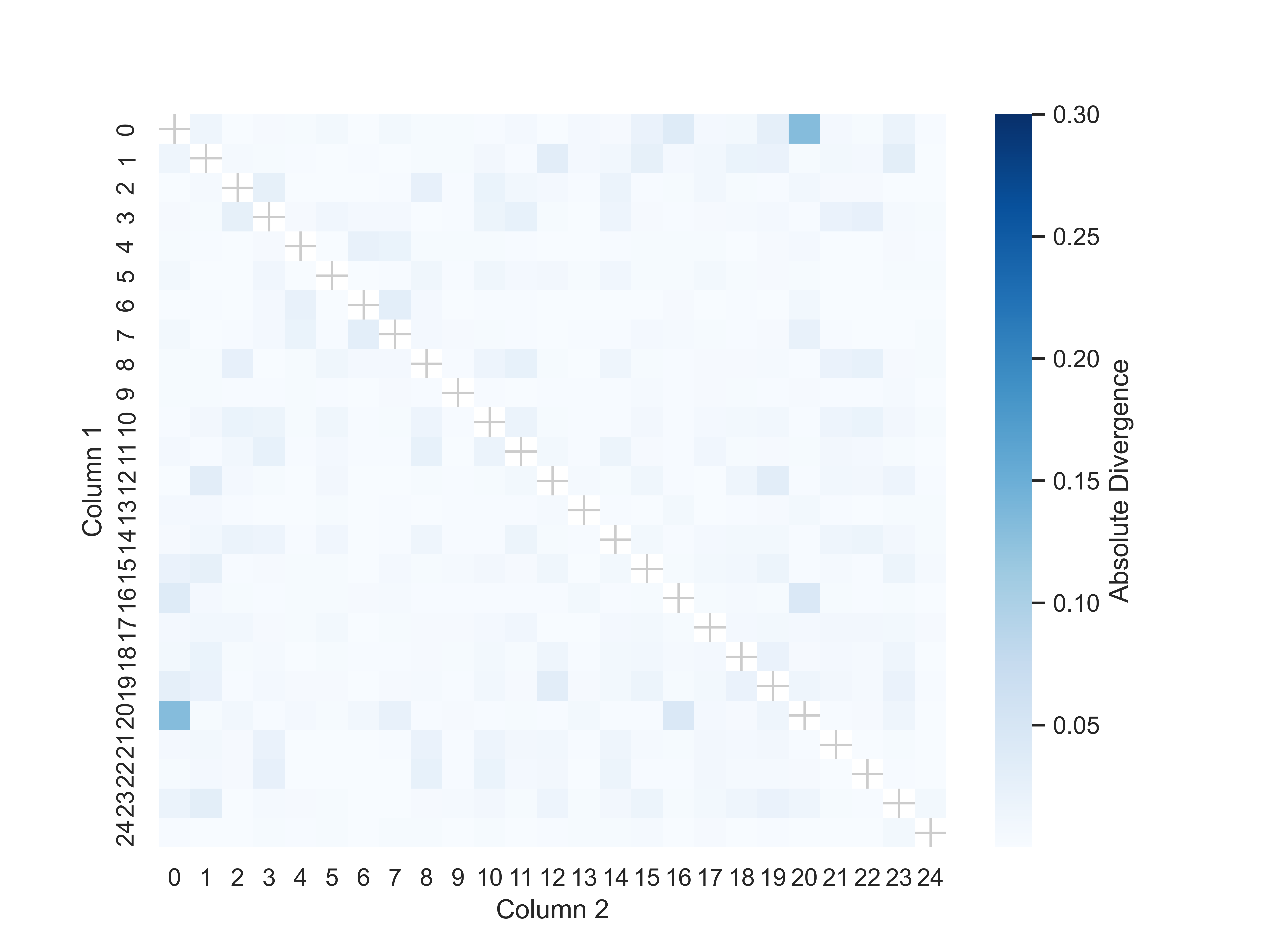}
        \caption{TabSyn}
    \end{subfigure} 
    \begin{subfigure}[t]{0.32\textwidth}
        \centering
        \includegraphics[width=\textwidth]{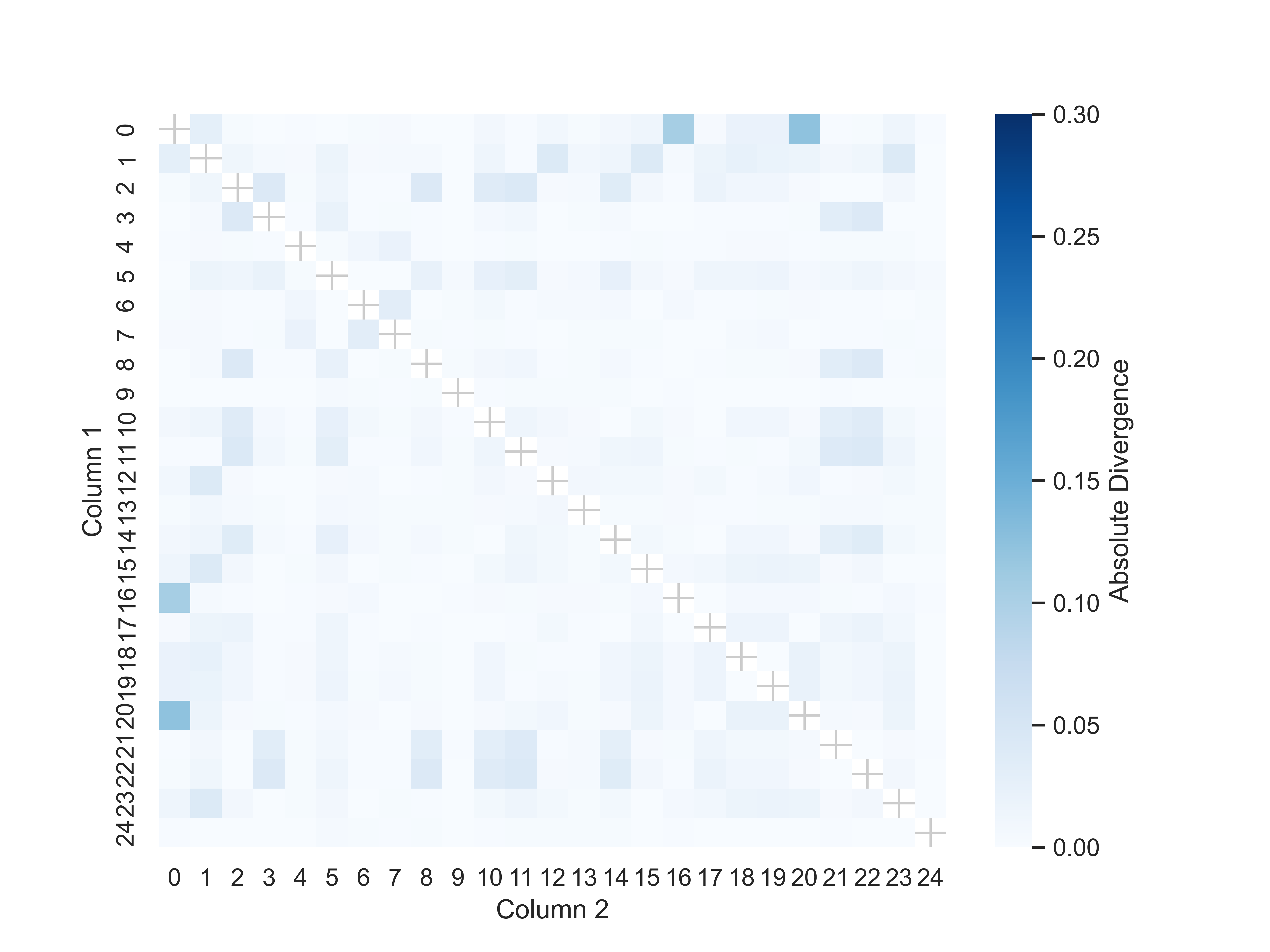}
        \caption{Ours (LLM-TabLogic)}
    \end{subfigure}
    
    \caption{Heatmap illustrating the absolute divergence in pairwise column correlations between the synthetic and real data.
    Lighter colors indicate smaller differences and better alignment. 
    SMOTE, TabSyn, and LLM-TabLogic exhibit the closest alignment with the real data, outperforming other methods.}
    \label{fig:heatmap}
\end{figure*}

    


Data fidelity is evaluated using a set of data accuracy and diversity metrics. 
The main objective of data fidelity evaluation is to assess whether the synthetic data is sufficiently authentic and accurate when compared to the real data.

For data accuracy, we first utilize low-order statistics, such as column-wise density estimation and pairwise correlation scores (i.e., Pearson Correlation), to measure the general statistical properties of the columns. 
As shown in \autoref{tab:accuracy}, SMOTE achieves the overall best performance on both datasets, followed {\it closely} by LLM-TabLogic and then TabSyn.
Secondly, for high-order statistics, which capture more complex statistical relationships, the 
$\alpha$-Precision scores demonstrate that LLM-TabLogic outperforms all other methods, achieving 99.15\% and 93.92\% on the Retailing and Purchasing datasets, respectively. TabSyn and SMOTE follow closely, with values of 98.48\% and 93.19\%.
To further examine the accuracy of synthetic data generated by various models, we analyzed the density distributions of  representative columns in the Retailing dataset.
First, we presneted the distribution of the three continuous columns---{\tt item profit ratio}, {\tt product price}, and {\tt latitude}---in \autoref{fig:accuracy1}. 
As shown in \autoref{fig:accuracy1}, TabSyn and LLM-TabLogic perform the overall best, producing distributions that are nearly {\it identical} to the real data.
While CTGAN and GReaT outperform than TabDDPM and generate distribution somewhat close to the real data, they exhibit noticeable gaps in peak regions, showing biases to the real data.
SMOTE presents lower variations in these regions, but it is overall slightly less accurate than TabSyn and LLM-TabLogic.
For categorical columns, such as {\tt shipping mode}, {\tt payment type}, and {\tt order status}, as shown in \autoref{fig:accuracy2}, LLM-TabLogic outperforms all other methods, accurately replicating category counts in these columns. 
TabSyn performs slightly worse than LLM-TabLogic but better than SMOTE. 
These results highlight LLM-TabLogic and TabSyn is superior to SMOTE for categorical data generation.
Additionally, we visualize the absolute divergence in pairwise column correlations between the real Retailing dataset and its synthetic counterpart using heatmaps, providing a clear and intuitive representation of the differences.
As presented in \autoref{fig:heatmap}, SMOTE, Tabyn, and LLM-TabLogic, though comparable to each other, achieve much better performance than other methods (e.g., CTGAN, TabDDPM, and GReaT) in terms of pairwise column correlation, with a {\it closer} match to the real data.

For the data diversity, we continue by evaluating how well the synthetic data captures all modes present in the real data using the average coverage score described in \autoref{eq:coverage_score}. 
The experimental results are presented in \autoref{tab:accuracy}.
As shown in \autoref{tab:accuracy}, for single columns, LLM-TabLogic and SMOTE achieve the highest values on the Purchasing and Retailing datasets, respectively. 

LLM-TabLogic, Tabsyn, and SMOTE nearly cover all modes for both categorical and numerical columns.
To assess the diversity of joint feature interactions, we used $\beta$-Recall scores. SMOTE, which explicitly balances minority classes, achieves the highest scores across both datasets, followed by LLM-TabLogic and GReaT, which rank second for each dataset, respectively. Notably, TabDDPM performs poorly on both accuracy and diversity metrics, indicating its inability to generate sufficiently diverse values.
In summary, LLM-TabLogic strikes an excellent balance between accuracy and diversity, outperforming other models in most aspects, particularly in synthesizing continuous and categorical data.

\subsection{Inter-Column Relationship Preservation}

\begin{table}[t]
\centering
\caption{Inter-column relationship identified in the Retail and Purchasing datasets.}
\label{tab:relationships}
\resizebox{\textwidth}{!}{%
\begin{tabular}{|c|ccc|c|}
\hline
\textbf{Dataset} &
  \textbf{Relationship} &
  \textbf{Group} &
  \textbf{Attributes} &
  \textbf{Condition} \\ \hline
\multicolumn{1}{|c|}{\multirow{10}{*}{\bf{Retailing}}} &
  \multirow{6}{*}{\begin{tabular}[c]{@{}c@{}}Hierarchical \\ Consistency\end{tabular}} &
  Geographical (Orders) &
  \begin{tabular}[c]{@{}c@{}}$x_{1,j}$ (order city), $x_{2,j}$ (order state), $x_{3,j}$ (order country), \\ $x_{4,j}$ (order region), $x_{5,j}$ (order market)\end{tabular} &
  $(x_{i,j})_{i \in G_1} \in C_{1,j}$
 \\ \cline{3-5} 
\multicolumn{1}{|c|}{} &
   &
  Geographical (Customers) &
  \begin{tabular}[c]{@{}c@{}}$x_{6,j}$ (customer city), $x_{7,j}$ (customer state), \\ $x_{8,j}$ (customer country)\end{tabular} &
  $(x_{i,j})_{i \in G_2} \in C_{2,j}$ \\ \cline{3-5} 
\multicolumn{1}{|c|}{} &
   &
  Products Information &
  \begin{tabular}[c]{@{}c@{}}$x_{9,j}$ (category ID), $x_{10,j}$ (category name), $x_{11,j}$ (department ID), \\ $x_{12,j}$ (department name), $x_{13,j}$ (product card ID), \\ $x_{14,j}$ (product category ID), $x_{15,j}$ (product name)\end{tabular} &
  $(x_{i,j})_{i \in G_3} \in C_{3,j}$ \\ \cline{2-5} 
\multicolumn{1}{|c|}{} & \begin{tabular}[c]{@{}c@{}}Temporal\\ Dependency\end{tabular} &
  Delivery Time &
  $x_{1,j}$ (order date), $x_{2,j}$ (delivery time) &
  $D_{1,j}: x_{1,j} < x_{2,j}$ \\ \cline{2-5} 
\multicolumn{1}{|c|}{} &
  \multirow{3}{*}{\begin{tabular}[c]{@{}c@{}}Mathematical\\ Dependency\end{tabular}} &
  \multirow{3}{*}{Sales Values} &
  \multirow{3}{*}{\begin{tabular}[c]{@{}c@{}}$x_{1,j}$ (quantity), $x_{2,j}$ (price), \\ $x_{3,j}$ (discount rate), $x_{4,j}$ (discount value), \\ $x_{5,j}$ (original price), $x_{6,j}$ (sales price)\end{tabular}} &
  $D_{2,j}: x_{5,j} = x_{1,j} \times x_{2,j}$ \\ \cline{5-5} 
\multicolumn{1}{|c|}{} &
   &
   &
   &
  $D_{3,j}: x_{5,j} = x_{1,j} \times x_{2,j} \times x_{3,j}$ \\ \cline{5-5} 
\multicolumn{1}{|c|}{} &
   &
   &
   &
  $D_{4,j}: x_{4,j} = x_{7,j} - x_{5,j}$ \\ \hline\hline
\multicolumn{1}{|c|}{\multirow{5}{*}{\bf{Purchasing}}} &
  \begin{tabular}[c]{@{}c@{}}Hierarchical \\ Consistency\end{tabular} &
  Supplier Information &
  $x_{1,j}$ (supplier number), $x_{2,j}$ (supplier name) &
  $(x_{i,j})_{i \in G_1} \in C_{1,j}$\\ \cline{2-5} 
\multicolumn{1}{|c|}{} &
  \multirow{2}{*}{\begin{tabular}[c]{@{}c@{}}Temporal\\ Dependency\end{tabular}} &
  \multirow{2}{*}{Delivery Time} &
  \multirow{2}{*}{\begin{tabular}[c]{@{}c@{}}$x_{3,j}$ (order created date), $x_{4,j}$ (planned delivery time), \\ $x_{5,j}$ (realistic delivery time)\end{tabular}} &
  $D_{1,j}: x_{3,j} < x_{4,j}$ \\ \cline{5-5} 
\multicolumn{1}{|c|}{} &
   &
   &
   &
  $D_{2,j}: x_{3,j} < x_{5,j}$ \\ \cline{2-5} 
\multicolumn{1}{|c|}{} &
  \multirow{2}{*}{\begin{tabular}[c]{@{}c@{}}Mathematical\\ Dependency\end{tabular}} &
  \multirow{2}{*}{Purchase Values} &
  \multirow{2}{*}{\begin{tabular}[c]{@{}c@{}}$x_{6,j}$ (quantity), $x_{7,j}$ (price), \\ $x_{8,j}$ (discount rate), $x_{9,j}$ (net amount), $x_{10,j}$ (gross amount)\end{tabular}} &
  $D_{3,j}: x_{9,j} = x_{6,j} \times x_{7,j}$ \\ \cline{5-5} 
\multicolumn{1}{|c|}{} &
   &
   &
   &
  $D_{4,j}: x_{10,j} = x_{9,j} \times (1-x_{8,j})$ \\ \hline
\end{tabular}%
}
\end{table}

\begin{table}[ht]
\centering
\caption{Evaluation results for preserving inter-column relationships in synthetic tabular data. 
Higher values indicate better performance. 
The best and second-best results are highlighted in {\color{blue}\bf blue} and {\bf black boldfaced}, respectively.}
\label{tab:logic}
\resizebox{0.9\textwidth}{!}{%
\begin{tabular}{cccccccc}
\toprule
\multirow{2}{*}{\textbf{Datasets}} &
\multirow{2}{*}{\textbf{Metrics}} &
  \multicolumn{1}{c}{\textbf{Interpolation}} &
  \multicolumn{3}{c}{\textbf{Latent Space-based}} &
  \textbf{PLM-based} & \textbf{Ours}\\ \cmidrule(lr){3-3} \cmidrule(lr){4-6} \cmidrule(lr){7-7} \cmidrule(lr){8-8} 
 &  &
  \textbf{SMOTE} &
  \textbf{CTGAN} &
  \textbf{TabDDPM} &
  \textbf{TabSyn} &
  \textbf{GReaT} &
  \textbf{LLM-TabFlow} \\ \midrule
\multirow{3}{*}{\bf{Retailing}} &
{HCS} &
  \textbf{98.09$\pm$0.11} &
   39.23$\pm$0.13 &
   16.07$\pm$0.21 &
   71.63$\pm$0.28 &
   98.01$\pm$0.31 &\textcolor{blue}{\textbf{99.90$\pm$0.10}} \\ 
& 
{MDI} &
   87.03$\pm$0.13 &
   38.87$\pm$0.25 &
   59.08$\pm$0.25 &
   68.34$\pm$0.38 &
   \textbf{97.37$\pm$0.27} &
  \textcolor{blue}{\textbf{99.95$\pm$0.05} }\\ 
& 
{DSI} &
   74.94$\pm$0.12 &
   30.55$\pm$0.13 &
   79.72$\pm$0.12 &
    66.94$\pm$0.21 &
   \bf{98.35$\pm$0.01} &
  \textcolor{blue}{\textbf{98.36$\pm$0.11}} \\\hline
\multirow{3}{*}{\bf{Purchasing}} &
{HCS} &
  81.17$\pm$0.29 &
   0.00$\pm$0.00 &
   0.01$\pm$0.05 &
   54.99$\pm$0.23 &
   \textbf{99.92$\pm$0.18 }&
  \textcolor{blue}{\textbf{99.89$\pm$0.11}} \\
& 
{MDI} &
   84.82$\pm$0.29 &
   36.69$\pm$0.36 &
   63.49$\pm$0.23 &
   56.30$\pm$0.41 &
   \textbf{94.54$\pm$0.36} &
  \textcolor{blue}{\textbf{99.92$\pm$0.08}} \\
& 
{DSI} &
   \textcolor{blue}{\bf{99.75$\pm$0.02}} &
   59.39$\pm$0.23 &
   96.62$\pm$0.21 &
   93.15$\pm$0.11 &
   94.45$\pm$0.32 &
 \textbf{96.83$\pm$0.12}\\\bottomrule

\end{tabular}%
}
\end{table}

Through LLM reasoning, the inter-column relationships in the two datasets are identified, as presented in \autoref{tab:relationships}.
These relationships are categorized into different types, with corresponding attributes and conditions specified for both datasets.
To quantify the inter-column consistency and dependency, we employ three evaluation metrics---HCS, MDI, and DSI---proposed in \citet{long2025evaluating}. 
The results are presented in \autoref{tab:logic}. 
As shown, LLM-TabLogic achieves a perfect value of 100\% in both consistency and dependency, demonstrating its ability to fully preserve the logical relationships within the real data. 
GReaT performs {\it slightly} worse, exhibiting strong inter-column relationship preservation but with some errors in capturing dependencies across multiple columns.
In contrast, latent space-based generative models (e.g., CTGAN, TabDDPM, and TabSyn) perform {\it significantly} worse, particularly in maintaining hierarchical consistency across both datasets. 
Interestingly, SMOTE outperforms most advanced generative models in preserving inter-column logical relationships.
To further assess dependency preservation, we evaluate the DSI score by analyzing previously unseen relationships between variables and target values. 
Specifically, we examine financial attributes to assess their impact on target outcomes across both datasets. 
LLM-TabLogic achieves the overall best performance, while SMOTE and GReaT perform well on individual datasets.
These results highlight LLM-TabLogic's strength in preserving both simple logical relationships and complex inter-column dependencies within tabular data, setting a new benchmark for synthetic tabular data generation with preserved inter-column relationships.

\subsection{Utility: Machine Learning Efficiency}
\label{sec:utility}

\begin{table}[ht]
\centering
\caption{Evaluation results of machine learning ({\bf classification}) efficiency in terms of the AUC score. 
Higher AUC values indicate better performance. 
The best and second-best results are highlighted in {\color{blue}\bf blue} and {\bf black boldfaced}, respectively.}
\label{tab:utility}
\resizebox{0.9\textwidth}{!}{%
\begin{tabular}{cccccccc}
\toprule
 &
  \multicolumn{1}{c}{\textbf{Real Dataset}} &
  \multicolumn{1}{c}{\textbf{Interpolation}} &
  \multicolumn{3}{c}{\textbf{Latent Space-based}} &
  \multicolumn{1}{c}{\textbf{PLM-based}} &
  
  \begin{tabular}[c]{@{}c@{}}\textbf{Ours}\end{tabular} \\ 
  \cmidrule(lr){2-2} \cmidrule(lr){3-3} \cmidrule(lr){4-6} \cmidrule(lr){7-7} \cmidrule(lr){8-8} 
  \textbf{Datasets} &
  \textbf{Real} &
  \textbf{SMOTE} &
  \textbf{CTGAN} &
  \textbf{TabDDPM} &
  \textbf{TabSyn} &
  \textbf{GReaT} &
  \textbf{LLM-TabFlow} \\ \hline
\bf{Retailing}  & 88.17 & 69.59±0.33 & 68.91±0.18 & 49.68±0.91 & \textbf{71.55±0.24} & 71.15±0.14 & \textcolor{blue}{\textbf{71.90±0.20}} \\ 
\hline
\bf{Purchasing} & 84.37 & \textcolor{blue}{\textbf{83.96±0.44}} & 53.31±0.60 & 49.72±1.28 & 62.26±1.38 & 50.06±0.75 & \textbf{63.45±0.92} \\\bottomrule
\end{tabular}%
}
\end{table}

While data utility, as illustrated in \autoref{fig:evaluation_framework}, consists of two sub-dimensions for evaluation, this experiment mainly focuses on quantitatively assessing the utility of the generated synthetic data through its machine learning efficiency.
The experimental results in \autoref{tab:utility} demonstrate that LLM-TabLogic achieves the overall best performance across both datasets, obtaining the highest AUC on the Retailing dataset and the second-highest on the Purchasing dataset.
Specifically, in the Retailing dataset, our approach outperforms TabSyn in terms of AUC scores for classification tasks. 
GReaT is just behind TabSyn but surpasses SMOTE, demonstrating the advantages of LLM-based training and sampling in generating synthetic tabular data.  
In contrast, GANs and TabDDPM models are {\it considerably} less effective in improving classification efficiency with large-scale synthetic data. 
However, all these methods still lag significantly behind those trained on the real dataset.
In the Purchasing dataset, SMOTE achieves highest AUC score and achieves performance close to that of real data. 
LLM-TabLogic, which achieves the second-best AUC, along with other methods, still lags behind real data, highlighting the ongoing challenge of achieving parity with real data in machine learning efficency, despite the advancements in generative modeling techniques.
Moreover, the ability of synthetic data to preserve inter-column relationships directly impacts its utility in real-world applications, such as process simulation and decision-making in finance and delivery of supply chains.
Based on the results in \autoref{tab:logic} and \autoref{tab:utility}, LLM-TabLogic generates synthetic tabular data that can effectively substitute real data, achieving {\it satisfactory} performance in real-world machine learning tasks.

\subsection{Privacy Preservation}\label{sec:privacy}

\begin{table}[h]
\centering
\caption{Evaluation results of privacy preservation in terms of the DCR and C2ST. 
Lower DCR values indicate better privacy preservation, with the best and smallest value being 50\%.
Higher C2ST values indicate better performance in terms of accuracy. 
The best and second-best results are highlighted in {\color{blue}\bf bold} and {\bf black boldfaced}, respectively.}
\label{tab:privacy}
\resizebox{0.9\textwidth}{!}{%
\begin{tabular}{cccccccc}
\toprule[1pt]
\multirow{2}{*}{\textbf{Datasets}} &
\multirow{2}{*}{\textbf{Metrics}} &
  \multicolumn{1}{c}{\textbf{Interpolation}} &
  \multicolumn{3}{c}{\textbf{Latent Space-based}} &
  \textbf{PLM-based} & \textbf{Ours}\\ \cmidrule(lr){3-3} \cmidrule(lr){4-6} \cmidrule(lr){7-7} \cmidrule(lr){8-8} 
 &  &
  \textbf{SMOTE} &
  \textbf{CTGAN} &
  \textbf{TabDDPM} &
  \textbf{TabSyn} &
  \textbf{GReaT} &
  \textbf{LLM-TabFlow} \\ \hline
\multirow{2}{*}{\textbf{Retailing}} &
DCR & 99.85±0.21 & \textbf{89.80±0.36} & \textcolor{blue}{\textbf{89.15±0.28}} & 90.18±0.27 & 90.19±0.25 & 90.03±0.23 \\ 
&
    C2ST &  \textcolor{blue}{\textbf{92.64±0.29}} & 17.65±0.32 & 0.00±0.00 & 52.89±0.23 & 38.83±0.16 & \textbf{66.75±0.15} \\ \hline
\multirow{2}{*}{\textbf{Purchasing}} &
DCR & 96.62±0.15 & \textbf{90.67±0.20}& 95.02±0.21 & 90.60±0.24 & 91.89±0.25 & \textcolor{blue}{\textbf{90.43±0.22}} \\ 
&
    C2ST &  \textcolor{blue}{\textbf{98.11±0.57}} & 66.92±0.69 & 0.27±0.13 & 93.93±0.39 & 76.34±0.37 & \textbf{95.38±0.30}\\
\bottomrule
\end{tabular}%
}
\end{table}

We further evaluate the proposed method alongside the five baseline approaches in the privacy dimension. 
As described in \autoref{sec:privacy}, DCR (see \autoref{eq:DCR})---a metric that quantifies the similarity between synthetic and real data---is used to evaluate privacy preservation. 
Additionally, since balancing privacy and utility is crucial for effective tabular data generation, we adopt the Classifier Two-Sample Test (C2ST) (see \autoref{sec:data_fidelity}) to measure the {\it distinguishability} of real and synthetic data.
For this experiment, a logistic regression classifier is used to compute the C2ST.
The results in \autoref{tab:privacy} show that LLM-TabLogic and TabDDPM achieve the best privacy-preserving performance (highest DCR values) on the Purchasing and Retailing datasets, respectively. 
However, TabDDPM obtains an extremely low C2ST value, indicating that its strong privacy preservation performance comes at the expensive cost of accuracy.
In contrast, SMOTE and LLM-TabLogic achieve the highest and second-highest C2ST scores on the Retailing and Purchasing dataset, respectively, demonstrating strong accuracy in data synthesis. 
While SMOTE shows the best accuracy in terms of accuracy (highest C2ST value), it achieves the worst (lowest DCR value) privacy preservation performance, highlighting its high accuracy cost at preserving privacy. 
Our method, however, provides the best balance between privacy and accuracy, outperforming both SMOTE and state-of-the-art generative models. 
This balance is crucial for applications that require both robust data privacy protection and high data utility, making our approach particularly well-suited for real-world industrial scenarios that demand both privacy guarantees and accurate data synthesis.

\section{Discussion and Implications}
\label{sec:discussion}

We propose a novel approach---LLM-TabLogic---that integrates LLM reasoning with score-based diffusion models to generate synthetic tabular data. 
This approach significantly outperforms traditional diffusion-based models in capturing complex inter-column logical relationships. Unlike conventional methods that embed all features with intricate relationships directly into latent space, LLM-TabLogic introduces an innovative framework that compresses complex tabular data into simplified representations for generation. These representations are subsequently decompressed back into their original form after sampling.
This approach not only enhances the model's ability to handle complex relationships but also provides a more efficient and interpretable framework for tabular data synthesis.
Experimental results demonstrate that PLM-based models excel at capturing  inter-column relationships, which may due to their autoregressive nature. 
However, these methods struggle to accurately preserve the low- and high-order statistics of the original (real) dataset and require approximately {\it ten} times more training and inference time compared to latent space-based models.
Additionally, it was observed that GReaT can generate unseen categories for product names, revealing the inherent uncertainty of LLM-based generative models.
This finding underscores the need for caution when using such models to generate complex categorical data.
In contrast, our approach mitigates the uncertainties, providing a more reliable method for tabular data generation.
It outperforms existing latent-space-based models in accurately handling datasets with diverse logical dependencies. 
Furthermore, LLM-TabLogic enhances computational efficiency and improves the capture of heterogeneous data distributions in latent space compared to other LLM-based models.
Overall, LLM-TabLogic achieves a much better balance between accuracy, utility, and privacy, positioning it as a robust solution for synthetic tabular data generation.

However, this work has some limitations.
Firstly, the identification of relationships in the data is influenced by the knowledge of the LLMs. 
There is no guarantee that LLMs can capture all complex relationships, particularly when column names are abbreviated or the descriptions are ambiguous or inaccurate. 
This reliance on LLMs may introduce biases or variances in understanding the underlying data structure.
Secondly, the experimental evaluation is limited to datasets within the supply chain domain. 
To ensure broader applicability, our method needs to be tested on more diverse and comprehensive tabular datasets from other industrial sectors, such as healthcare, finance, and social sciences. 
This would help verify the generalizaubility of LLM-TabLogic across different types of domain knowledge and data structures.
Additionally, privacy concerns arise if column names and descriptions contain sensitive information, as LLMs process these inputs to infer relationships. 
This could potentially expose private information, especially in domains such as healthcare or finance.

Despite these limitations, LLM-TabLogic shows significant potential for real-world uses across various industries. 
Beyond preserving data privacy and enhancing machine learning tasks, it is, to the best of our knowledge, the {\it first} framework capable of capturing and replicating complex real-world dependencies in synthetic tabular data. 
This capability is particularly valuable for conducting simulations in digital twins \citep{xu2025multi}, enabling more accurate business analysis, predictive modeling, and data-driven decision-making.
For example, LLM-TabLogic can be applied to model complex relational data across multiple competing business entities, addressing the data scarcity problems in industrial domains caused by privacy and reluctance to share data \citep{xu2025multi,almahri2024enhancing}. 
In agent-based simulations, LLM-TabLogic would generates synthetic data that accurately reflects the interactions between agents, enabling researchers to study complex systems such as supply chains \citep{xu2024implementing, xu2025multi}, financial markets \citep{proselkov2024financial}, and social networks. 
Similarly, high-quality, realistic synthetic data would be used to train and fine-tune autonomous agents for customized scenarios, enhancing their capabilities in dynamic environments, such as managing supply chain operations and making sophisticated decision making in autonomous supply chains \citep{xu2024multi,xu2024implementing,xu2024towards}. 
These industrial implications highlight LLM-TabLogic's significant impact in generating high-fidelity, privacy-preserved synthetic data, positioning it as a transformative tool for driving data-driven innovation in both industry and academia.

\section{Conclusion and Future Work}
\label{sec:conclusion}

This paper addresses the challenge of preserving inter-column relationships in synthetic tabular data generation for real-world applications. 
We propose an evaluation framework comprising six dimensions to evaluate the performance of synthetic tabular data in industrial contexts, where both high utility and strong privacy preservation are essential.
We have proposed LLM-TabLogic, a novel synthetic tabular data generation approach that integrates LLM reasoning with latent space-based data generation.
Additionally, we conduct extensive experiments to evaluate the performance of the proposed method and the other five baselines under the proposed evaluation framework. 
These experiments show that existing methods struggle to preserve logical inter-column relationships, often failing to maintain dependencies and consistenc. 
In contrast, effectively upholds hierarchical consistency and logical dependencies while achieving the best trade-off among data fidelity, privacy, and utility.

To the best of our knowledge, LLM-TabLogic is the first solution capable of synthesizing tabular data while simultaneously preserving inter-column relationships and ensuring privacy. 
This highlights its potential for generating high-quality data for industrial applications, including simulation and advanced decision-making.
Moreover, this approach lays the foundation for integrating LLM reasoning with tabular data generation, offering a pipeline that can be extended to other domains such as finance and healthcare. 
By doing so, LLM-TabLogic unlocks new opportunities for both research and real-world applications, fostering innovation in data-driven industries.

However, our study has certain limitations, as discussed in \autoref{sec:discussion}.
Addressing these limitations will be a key focus of our future work.
First, we will evaluate LLM-TabLogic on a broader range of diverse and challenging datasets, including those from various industrial domains and datasets with abbreviated or ambiguous column names and descriptions.
This will help assess the model’s adaptability and its ability to capture complex inter-column relationships across various contexts.
Second, we will explore to incorporate structural constraints to achieve more precise and context-aware data synthesis.
Another key focus of our future work is evaluating LLM-TabLogic's ability to identify inter-column relationships when sensitive information in column names or descriptions is anonymized.
Furthermore, while LLM-TabLogic effectively preserves logical relationships in complex tabular data, real-world tables often contain domain-specific constraints that are not always apparent, such as environmental limits (e.g., industrial temperature ranges) and natural rules (e.g., age must be over 18 for adulthood).
Incorporating new prompting methods into our model is crucial for maintaining these structured constraints, ensuring more precise and context-aware data synthesis.
Additionally, preserving column relationships across multiple tables is essential for generating realistic synthetic tabular data, especially in scenarios involving foreign key dependencies. Our future work will focus on identifying and maintaining these cross-table relationships to ensure consistency, which is critical for information systems in industries such as supply chain management, e-commerce, and healthcare.

\section{Data Availability Statement}

The datasets used in this study include both public and restricted-access sources. 
The UCI Adult Income Dataset~\citep{adult} and the DataCo Supply Chain Dataset~\citep{dataco} are publicly available. 
The MIMIC-III Clinical Database~\citep{mimic} requires credentialed access via PhysioNet due to privacy constraints. 
The Purchasing Dataset is proprietary and cannot be shared. 
Public datasets are cited in the references.

\appendix
\newpage


{\small
\bibliographystyle{plainnat}
\bibliography{reference}
}
\end{document}